\documentclass[acmtog, nonacm]{acmart}

\setcopyright{none}
\renewcommand\footnotetextcopyrightpermission[1]{}

\AtBeginDocument{%
  \providecommand\BibTeX{{%
    \normalfont B\kern-0.5em{\scshape i\kern-0.25em b}\kern-0.8em\TeX}}}

\definecolor{amber}{rgb}{1.0, 0.3, 0}
\definecolor{green}{rgb}{0, 0.7, 0}

\newcommand{\rv}[1]{{\color{black}{#1}}}

\newcommand{\reffig}[1]{{Fig.~\ref{fig:#1}}}
\newcommand{\reftab}[1]{{Tab.~\ref{tab:#1}}}
\newcommand{\refsec}[1]{{Sec.~\ref{sec:#1}}}
\newcommand{\refeqn}[1]{{Eqn.~(\ref{eqn:#1})}}

\newcommand{\eg}[1]{{\textit{e.g.,~}}}
\newcommand{\ie}[1]{{\textit{i.e.,~}}}

\usepackage{booktabs}
\usepackage[ruled]{algorithm2e} 
\usepackage{graphicx}
\usepackage{float}
\usepackage{stfloats}
\usepackage{amsmath}
\usepackage{makecell}
\usepackage{comment}
\usepackage{subcaption}
\usepackage{hyperref}

\SetAlFnt{\small}
\SetAlCapFnt{\small}
\SetAlCapNameFnt{\small}
\SetAlCapHSkip{0pt}

\begin{document}

\title{UniColor: A Unified Framework for Multi-Modal Colorization with Transformer}

\author{Zhitong Huang}
\email{luckyhzt@gmail.com}
\affiliation{
    \institution{City University of Hong Kong}
    \city{Hong Kong SAR}
    \country{China}
}
\authornote{\label{equal}Both authors contributed equally to this research.}

\author{Nanxuan Zhao}
\email{nanxuanzhao@gmail.com}
\affiliation{
    \institution{University of Bath}
    \city{Bath}
    \country{United Kingdom}
}
\authornotemark[1]

\author{Jing Liao}
\email{jingliao@cityu.edu.hk}
\affiliation{
    \institution{City University of Hong Kong}
    \city{Hong Kong SAR}
    \country{China}
}
\authornote{Corresponding author.}





\begin{abstract}
\par
We propose the first unified framework \textit{UniColor} to support colorization in multiple modalities, including both unconditional and conditional ones, such as stroke, exemplar, text, and even a mix of them. Rather than learning a separate model for each type of condition, we introduce a two-stage colorization framework for incorporating various conditions into a single model. In the first stage, multi-modal conditions are converted into a common representation of hint points. Particularly, we propose a novel CLIP-based method to convert the text to hint points. In the second stage, we propose a Transformer-based network composed of \textit{Chroma-VQGAN} and \textit{Hybrid-Transformer} to generate diverse and high-quality colorization results conditioned on hint points. Both qualitative and quantitative comparisons demonstrate that our method outperforms state-of-the-art methods in every control modality and further enables multi-modal colorization that was not feasible before. Moreover, we design an interactive interface showing the effectiveness of our unified framework in practical usage, including automatic colorization, hybrid-control colorization, local recolorization, and iterative color editing. Our code and models are available at \textit{\textcolor{blue}{ \url{https://luckyhzt.github.io/unicolor}}}.

\end{abstract}



\begin{teaserfigure}
    \centering
    \includegraphics[width=1.0\textwidth]{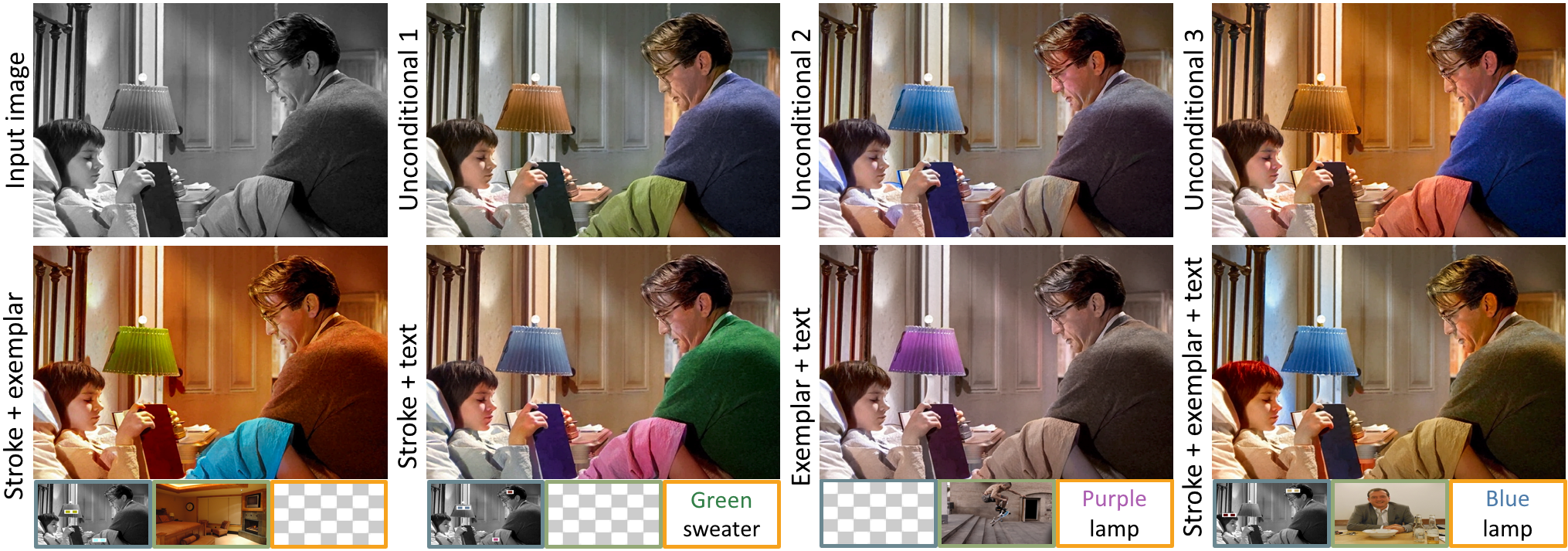}
    \caption{Given an input grayscale image, our unified framework UniColor is able to: (a) produce diverse colorization results unconditionally ($1^{st}$ row), and (b) colorize the image from a hybrid of controls ($2^{nd}$ row), including stroke (in blue frame), exemplar (in green frame), and text (in orange frame). Input image: Gregory Peck \& Mary Badham in film \textit{To Kill a Mockingbird}, 1962. Reference images are from MSCOCO.}
    \label{fig:teaser}
\end{teaserfigure}

\maketitle

\section{Introduction}

Colorization, a task of adding colors to grayscale images, has been actively studied recently~\cite{coltran, vivid-colorization, focus-person, chromagan}.  As an ill-posed problem, colorization often has a one-to-many mapping, where many results could be semantically meaningful and visually pleasing for a single input grayscale image. For example, the lamp and sweater in \reffig{teaser} could have many diverse colors in the colorized images. Previous unconditional colorization methods~\cite{colorful, letbecolor, chromagan, instance-colorization} can only generate a deterministic result for a single grayscale image, failed to maintain the diverse and expressive nature.
\rv{Some other methods \cite{diverse-vae, coltran, palette} can produce diverse colorization results, but fail to customize the colors based on user control.
To alleviate these problems}, various conditional methods have been proposed, which can be categorized into: stroke-based methods \cite{deep-prop, user-guided, interactive-col}, exemplar-based methods~\cite{exemplar-image, exemplar-style, exemplar-video}, and text-based methods~\cite{learn-color-language}. The generated colors depend on how these conditions are designed.

Although conditional methods allow customized results, they are still limited to a single modality, decreasing the flexibility of the general usage. In practice, a combination of different interaction manners is often required to achieve a satisfactory colorization result. For example, in the last result of \reffig{teaser}, we use different modalities to control different objects, \ie~exemplar for large object (sweater) and background (door), text for smaller object of the lamp, and stroke for details of the hair. Thus, a framework that enables multi-modal conditions for colorization is a natural choice and in demand for practical usage. In this work, we aim to develop the first unified colorization framework, producing diverse results under multi-modal controls, including both unconditional and conditional ones (\eg~stroke, exemplar, and text).

However, creating such a unified colorization framework is not an easy task with two unique challenges. 1). \textit{Multi-modal controls}: existing works often design and train the model supporting only a single type of user interaction and cannot generalize to the other ones directly. Given the complete different distributions on various conditions (\eg~stroke, exemplar, and text), how to encode different modalities and take control of results with freely integrated conditions is critical to the framework design. 2). \textit{Diversity and quality}: being a stochastic task, the framework needs to output diverse results with high quality, which are both semantically meaningful and visually pleasing.

To this end, we introduce \textit{UniColor}, a novel unified framework for colorization with multi-modal interactions. Our framework can colorize a grayscale image from scratch or based on conditions either in a single type or multi-modal ones. We mainly consider modalities, including stroke, exemplar, and text, which are the common interactive ways for the colorization task. To unify different modalities, we adopt a two-stage framework by taking hint points as an intermediate representation. A hint point is a point with conditional colors, where the minimum size is a pixel. This is because hint points can be naturally decomposed or extracted from different modalities. For the stroke-based condition, the hint points can be sampled along with strokes. For the exemplar-based condition, a colorful exemplar is warped to the input grayscale image based on semantic matching, and then hint points with high matching confidence can be selected. For the text-based condition, we introduce a new method based on CLIP embedding~\cite{clip} for assigning hint points on the objects corresponding to the input text. We thus convert all modalities into a unified representation (\ie~hint points) in the first stage. Then the model can concentrate on learning colorization in the second stage.

To ensure the diversity and quality of generated results, in the second stage, we take advantage of Transformer architecture~\cite{self-attention}, which has shown great success on various generation tasks~\cite{taming, dalle, ufc-bert, nvwa}. Given a grayscale image, together with hint points, we design a Transformer-based network for diverse colorization. More specifically, we first introduce \textit{Chroma-VQGAN}, a subnetwork composed of two separated gray and color encoders and one joint decoder, to disentangle chroma representation from the gray one. The chroma representation is discretized through a learned codebook, while gray representation remains continuous features for keeping input details. We then propose a \textit{Hybrid-Transformer} for predicting chrominance values. Different from previous works~\cite{taming, ufc-bert, nvwa} taking all the input and conditions as discrete tokens, our Hybrid-Transformer is created for mix-type inputs, including hint points in pure colors, continuous gray representation, and quantized chroma representation, which avoids quantization loss on the gray input and hint points. We follow a BERT-style~\cite{bert} training scheme for incorporating color hints in any position.

The extensive experiments on both unconditional and various conditional colorization tasks demonstrate the effectiveness of our method for generating diverse results with high-quality. We also design a UI tool for interactive colorization under our unified framework. Please see more details in the supplementary video demo. In summary, we make the following contributions:
\begin{itemize}
    \item We propose the first unified framework (UniColor) for interactive colorization, allowing multi-modal conditions in hybrid-mode.
    \item We present a method for unifying multi-modal conditions, including stroke, exemplar, and text, by taking hint points as an intermediate representation. Especially, we introduce a novel CLIP-based method for text-to-hint-point conversion. 
    \item We propose a colorization network composed of Chroma-VQGAN and Hybrid-Transformer for generating diverse and high-quality colorization results both conditionally and unconditionally.
    \item We design an interface, showing the effectiveness of our unified framework in practical usage.
\end{itemize}

\begin{figure*}[t]
	\centerline{\includegraphics[width=1.0\textwidth]{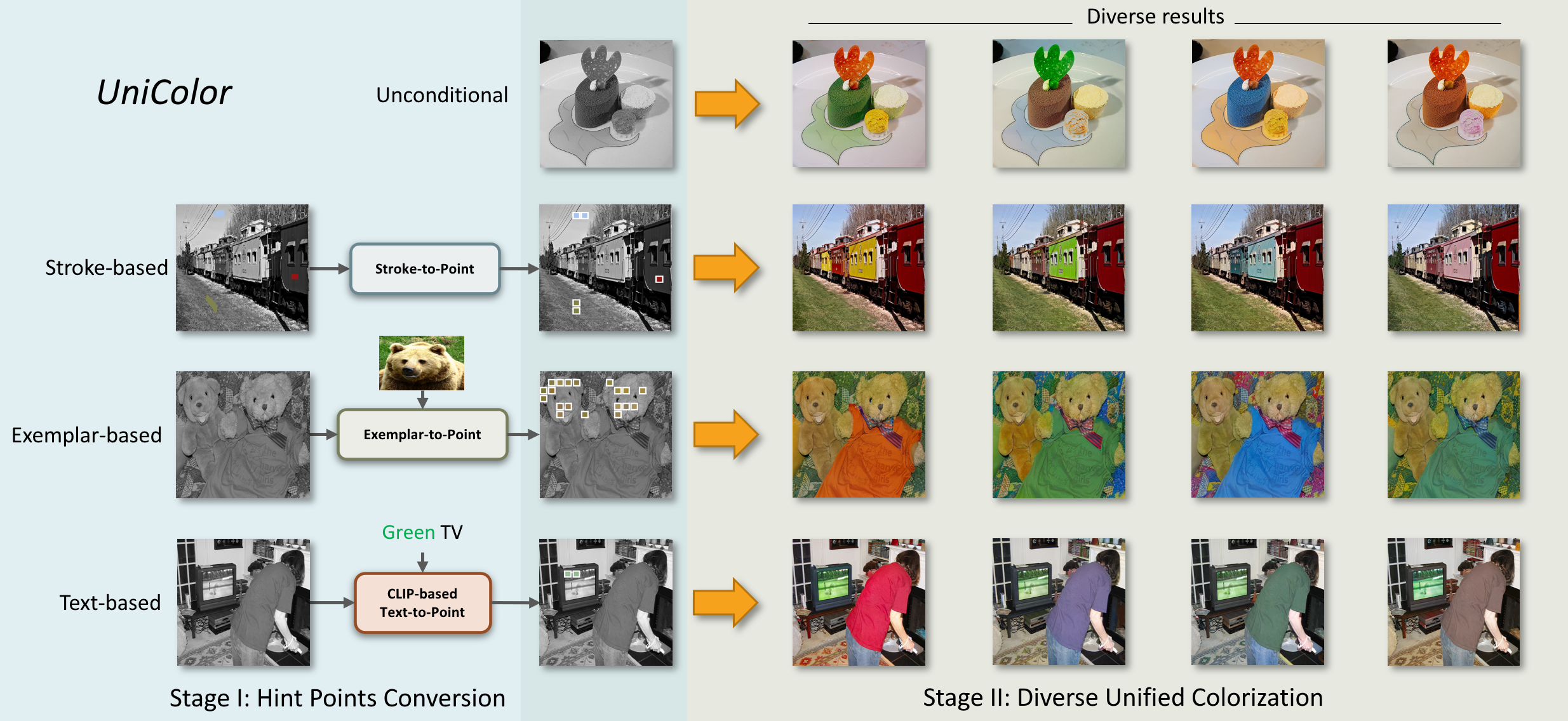}}
	\caption{Our unified colorization pipeline. The pipeline consists of two stages. In the first stage, all different conditions are unified as hint points. In the second stage, diverse results are generated automatically either from scratch or based on the condition of hint points. \rv{Input images: the $1^{st}$ row is from ImageNet and all others are from MSCOCO.} }
	\label{fig:s4_unified}
\end{figure*}

\section{Related Works}

\subsection{Unconditional Colorization}

Deep neural network has been proven to be successful in image colorization, where the model is learned from hundreds of thousands of gray-color image pairs. An earlier work~\cite{colorful} treats colorization as a classification task, and proposes an end-to-end network for learning the mapping from the grayscale images to the distribution of quantized chrominance values. Global semantic information is further added to guide the colorization~\cite{letbecolor}, where the model jointly learns to colorize and predict class labels.With the development of generative adversarial network (GAN)~\cite{GAN}, it has been used in colorization tasks~\cite{chromagan, image2image, image-color-gan}. To enhance the quality of colorization on images with multiple objects, instance-aware colorization is introduced~\cite{instance-colorization} by extracting object-level features. 

\rv{
\textbf{Diverse colorization.} As colorization is an ill-posed problem with multiple possible solutions, it is essential to generate diverse colorization results instead of a deterministic one. Cao et al.~\shortcite{diverse-gan} initialize the color channels with random noise and train the generator to produce diverse colors from the noise with adversarial loss. Deshpande et al.~\shortcite{diverse-vae} model the VAE-encoded color embeddings as a mixture of Gaussian distributions conditioned on the gray image, and sample diverse colors from the Gaussian distributions. A Gaussian conditional random field layer is further applied to the output color distribution~\cite{control-diverse}, to enhance global structural consistency and enable controllability in form of sparse color points. Wu et al.~\shortcite{vivid-colorization} propose a framework for diverse colorization results, with guidance from the diverse color priors generated by a pretrained GAN. Coltran~\cite{coltran} models the probability of color distribution with Transformer and samples diverse colors autoregressively. Recently, Palette~\cite{palette} trains a diffusion model~\cite{ddpm} to denoise the color images from the Gaussian noises, conditioned on the gray images, for diverse colorization.
}

\subsection{Conditional Colorization}

To allow user controls, some conditional colorization methods were proposed. According to the control modality, they can be categorized into: stroke-based colorization, exemplar-based colorization, and text-based colorization. However, there is no existing method unifying multi-modal controls.

\textbf{Stroke-based colorization.}
Stroke-based colorization requires users to draw color strokes as the condition. Earlier works~\cite{optim-color,natural-color} rely on optimization methods for propagating the user input strokes to the whole image, based on intensity and spatial position. The later work~\cite{deep-prop} learns the similarity mapping explicitly through a neural network by generating a probability map. Recently, rather than guiding with explicit similarity metrics, several methods~\cite{user-guided,interactive-col} propose to conduct end-to-end training for direct propagation.

\textbf{Exemplar-based colorization.} 
Exemplar-based colorization aims to colorize a grayscale image based on a user input reference image. The colorization pipeline usually consists of two steps. The first step warps the reference image to the grayscale input, based on either the deep similarity metrics~\cite{exemplar-image, exemplar-video} or transfer color from the reference image through a learned network~\cite{exemplar-style}. The second step is to generate the final colorization based on the warped reference. The performance of exemplar-based colorization largely relies on the fully warped or transferred reference, and it fails to generate feasible results in the regions with incorrect correspondences.

\textbf{Text-based colorization.}
Text-based colorization generates results according to the user-input text description, which often contains an object with a color word. Manjunatha et al.~\shortcite{learn-color-language} encodes the input text by an LSTM and fuses it with the visual feature of the input grayscale image through feature-wise affine transformations. Chen et al.~\shortcite{language-editing} introduces a recurrent attentive model to fuse the text feature with the image feature, but limits images to specific classes, such as flowers and geometry shapes. 
\rv{Bahng et al.~\shortcite{text_palette} generate color palette from text to control the global color distribution.}
Different from stroke-based and exemplar-based methods, text-based colorization is still in an early stage and needs further exploration.

\subsection{Transformer in Image Generation and colorization}

Transformer architecture~\cite{self-attention} is originally introduced for natural language processing, which models long-range relations between the input tokens through attention~\cite{structured-attention,joint-nmt}. It is then extended to the image domain by treating each image pixel as a visual word~\cite{igpt}. However, this direct usage cannot deal with high-resolution images as the computational cost increases quadratically with the number of tokens increasing. Later variants are introduced to improve the efficiency including: 1) restricting attention to local fields~\cite{image-transformer,ar-video}; 2) replacing the full attention with successive partial attentions~\cite{sparse-transformer, axial-transformer}; and 3) reducing the length of input tokens~\cite{vit, taming}. A two-stage framework for high-resolution image synthesis~\cite{taming} is proposed by using a convolutional neural network (CNN) based VQGAN to encode and tokenize the input image. This method not only reduces the length of input tokens but also incorporates the advantages of CNN (\eg~local interactions and inductive bias).

\textbf{Colorization with Transformers.} Transformer architecture is first introduced to image colorization in Coltran~\cite{coltran}. The colorization process is divided into three coarse-to-fine modules including a conditional autoregressive (AR) transformer, a color upsampler, and a spatial upsampler. The work regards each image pixel as a token and uses axial attention~\cite{axial-transformer} to enable image colorization in a resolution of 256$\times$256, but hard to be applied to higher resolution. The encoded feature of the grayscale image is used to modulate the layers of the transformer, which generates diverse colorized images under multinomial sampling. However, this method does not accept any kinds of controls and only supports unconditional colorization. Additionally, with pure Transformer architecture, the method cannot take advantage of the local interactions and inductive bias from CNN.

\section{UniColor}

We aim to design a unified colorization framework producing diverse results unconditionally or conditioned on a mixed set of multi-modal inputs, including stroke, exemplar, and text. To implement such a framework, we mainly face two challenges.

First, conditions under different modalities have various representations, which require different network architecture. For example, an exemplar image is usually warped by a correspondence network~\cite{exemplar-video} while the text needs to be encoded with the LSTM ~\cite{learn-color-language}. Therefore, multi-modal conditions are difficult to be incorporated into a single network. To deal with multi-modal conditions, we introduce hint points as an intermediate unified representation for all three types of conditions (\ie~stroke, exemplar, and text). As shown in  \reffig{s4_unified} (stage 1), all modalities are converted into the form of hint points, which are then used to guide and control the colorization process. This design also enhances the generalization of our framework since this representation can be easily extended to process new modalities.

Second, as a one-to-many mapping problem, the framework should generate colorization in diverse results, which are both aesthetically pleasing and semantically valid. To achieve this goal, we propose a network combining both VQGAN and Transformer architectures, which have shown promising results on general image generation and editing tasks~\cite{taming, ufc-bert, nvwa}. There are two benefits: 1) by quantizing the image into a codebook through VQGAN, the method can be converted into a classification formulation for sampling different results based on probabilities, and thus increases the diversity; 2) by building correlations among tokens in Transformer, the model learns global consistency for ensuring the quality of results. While naive usage of VQGAN and Transformer generates serious artifacts, we introduce novel Chroma-VQGAN and Hybrid-Transformer for our unique unified colorization problem.

An overview of our framework is shown in~\reffig{s4_unified}, which consists of two stages. The first stage called \textit{Hint Points Conversion} aims to convert multi-modal conditions (\ie~stroke $s$, reference image $I_r$, and text $t$) into a uniform hint point representation $h_c$. The second stage, called \textit{Diverse Unified Colorization}, aims to generate diverse high-quality colorization results from the input grayscale image~$I_g \in \mathbb{R}^{H\times W}$, conditioned on a mixed set of conditions $\mathbb{P}(\{s, I_r, t\})$ ($\mathbb{P}$ is the power set):
\begin{equation}
    \{\hat{I}_c^i\} \sim P(\hat{I}_c | I_g, \mathbb{H}(\mathbb{P}(\{s, I_r, t\})))
\end{equation}
where $\mathbb{H}$ is the Hint Points Conversion and $\hat{I}_c^i$ is the $i$-th image, sampled from the multinomial probability distribution $P$, in the set of diverse colorization results $\{\hat{I}_c^i\}$.

\subsection{Hint Points Conversion}  \label{sec:hint_point}

Hint points, a set of points assigned with target colors, as shown in~\reffig{s4_unified}, is an accurate and flexible way of representing the color condition. Different modalities commonly used in the colorization task (\ie~stroke~\cite{user-guided, interactive-col}, exemplar~\cite{exemplar-image, exemplar-video, exemplar-style}, and text~\cite{learn-color-language}) can be converted to hint points by different modules easily and then be used in a mixed way for practical application. For fast processing, we divide an image into a grid of cells with size $d\times d$. So a hint point corresponds to one cell in a single color.

\textbf{Stroke to Hint Points Conversion.} Given user-drawn strokes on a grayscale image, we traverse the cells along the stroke, and regard the cell as a hint point if the number of colored pixels within a cell surpasses a threshold (\eg~$0.75d$), then we assign the color of this hint point as the stroke's color. After repeating this process for all the strokes, the grayscale image with hint points will be sent into the second stage as shown in \reffig{s4_unified}.

\textbf{Exemplar to Hint Points Conversion.} The exemplar used in colorization is often an image sharing similar semantic content with the target one. To convert the exemplar image into hint points, we are inspired by the conventional way of warping the exemplar image to the grayscale image with semantic matching~\cite{exemplar-image, exemplar-video}. But different from previous works warping all the pixels equally, we only keep the $d\times d$ cells with high matching confidences as hint points. This is to avoid the injected noises caused by mismatches and allow more diverse sampling on final results. The confidences are measured from the output correlation matrix of the correspondence network in~\cite{exemplar-video}, which is based on pretrained VGG19~\cite{vggnet}. More specifically, we keep a hint point if the average similarity of its corresponding cell in the warped image is larger than a threshold (\ie, we set to 0.23 empirically), and assign the color to the hint point as the mean color of the corresponding cell.

\begin{figure}[t]
	\centerline{\includegraphics[width=1.0\linewidth]{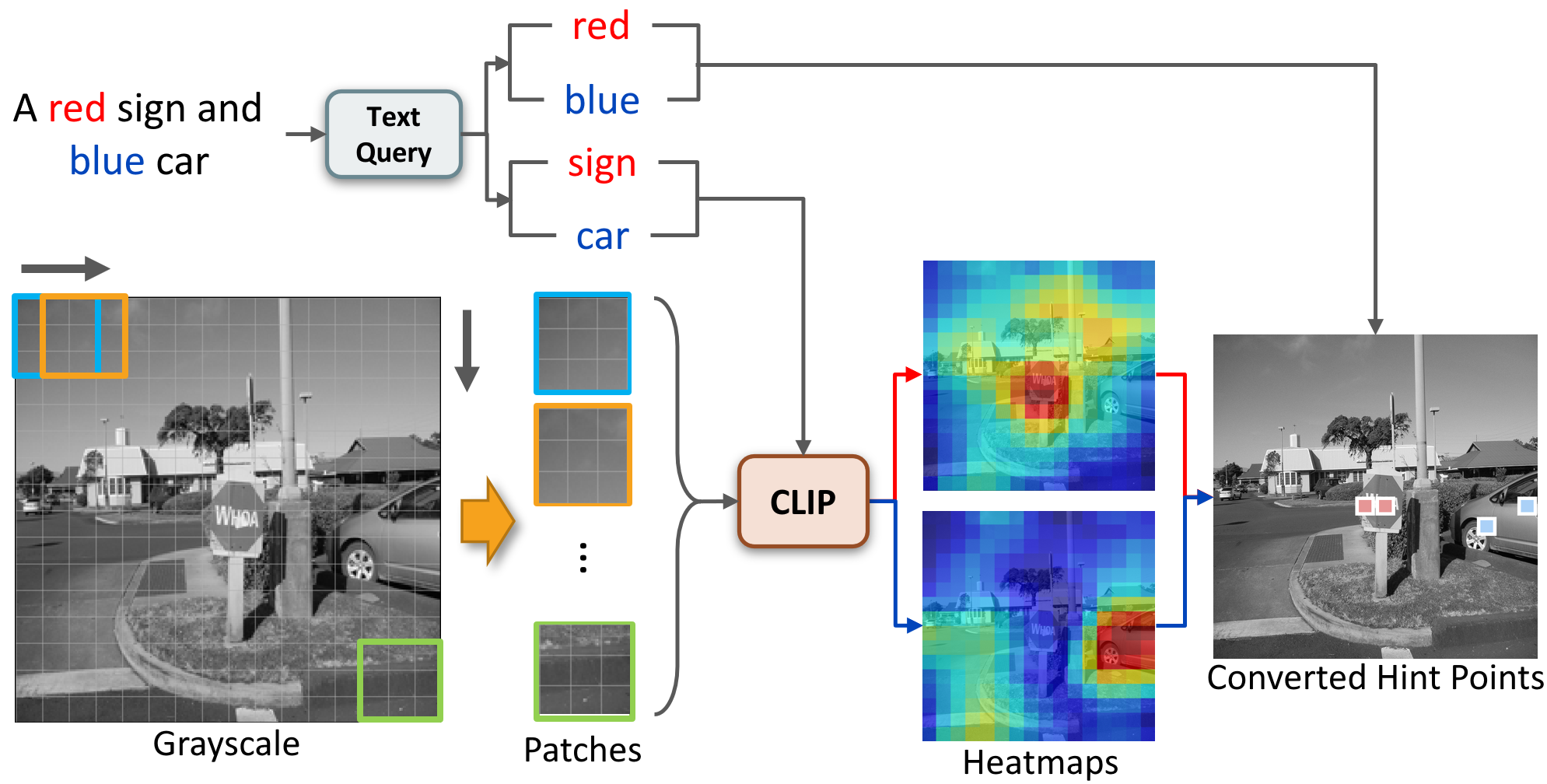}}
	\caption{Text-to-hint-points conversion. For each object concept, we calculate a correspondence map by measuring the similarity between cells and the textual concept through CLIP embedding, and the cells with top-2 correspondence values are selected as hint points. \rv{Input image: from MSCOCO.}}
	\label{fig:s4_text_based}
\end{figure}

\textbf{Text to Hint Points Conversion.} Text-based image generation and manipulation tasks have been actively studied recently as the development of natural language processing techniques. It is extended to colorization tasks but is still under-explored without a mature and conventional method. In view of this, we propose a novel method based on the Contrastive Language-Image Pre-training (CLIP) embedding~\cite{clip}, which has shown to be a powerful text-image representation. 
\rv{Note that compared with previous referring object segmentation methods~\cite{mcn, cris}, our CLIP-based method 
is training-free without the need for ground-truth annotations (\eg~RefCOCO~\cite{refcoco}) and can naturally deal with objects in open vocabulary.}
Given a text containing the described objects with colors, we first extract the object and color concepts based on the text parsing. We then divide the grayscale image into grid with a cell size of $d$, and slide an $n\times n$ window (\ie~we set $n=3$) with stride $1$ for extracting patches across the image. At each location, we calculate the correspondence value between the patch of $n\times n$ cells and each of the object concepts by cosine similarity between the features from the CLIP embedding, as shown in~\reffig{s4_text_based}, and this correspondence value contributes to all the $n\times n$ cells within a patch. For each of the object concepts, after averaging all values within a cell, we obtain a correspondence map, and regard the top-2 cells as hint points. The specified color of the corresponding object in the given text is further assigned to the hint points, according to a color table. 

\subsection{Diverse Unified Colorization}

Given color hints and a grayscale image, the second stage of UniColor aims to propagate the hint colors for generating diverse colorization results. We thus propose a Transformer-based architecture. As shown in~\reffig{structure}, it contains two sub-networks, one is the Chroma-VQGAN for learning a disentangled and quantized chroma representation from the continuous gray one, and the other is the Hybrid-Transformer for learning colorization based on unified condition.

\subsubsection{Chroma-VQGAN}  \label{sec:s4_chromavqgan}

\begin{figure*}[t]
	\centerline{\includegraphics[width=1.0\textwidth]{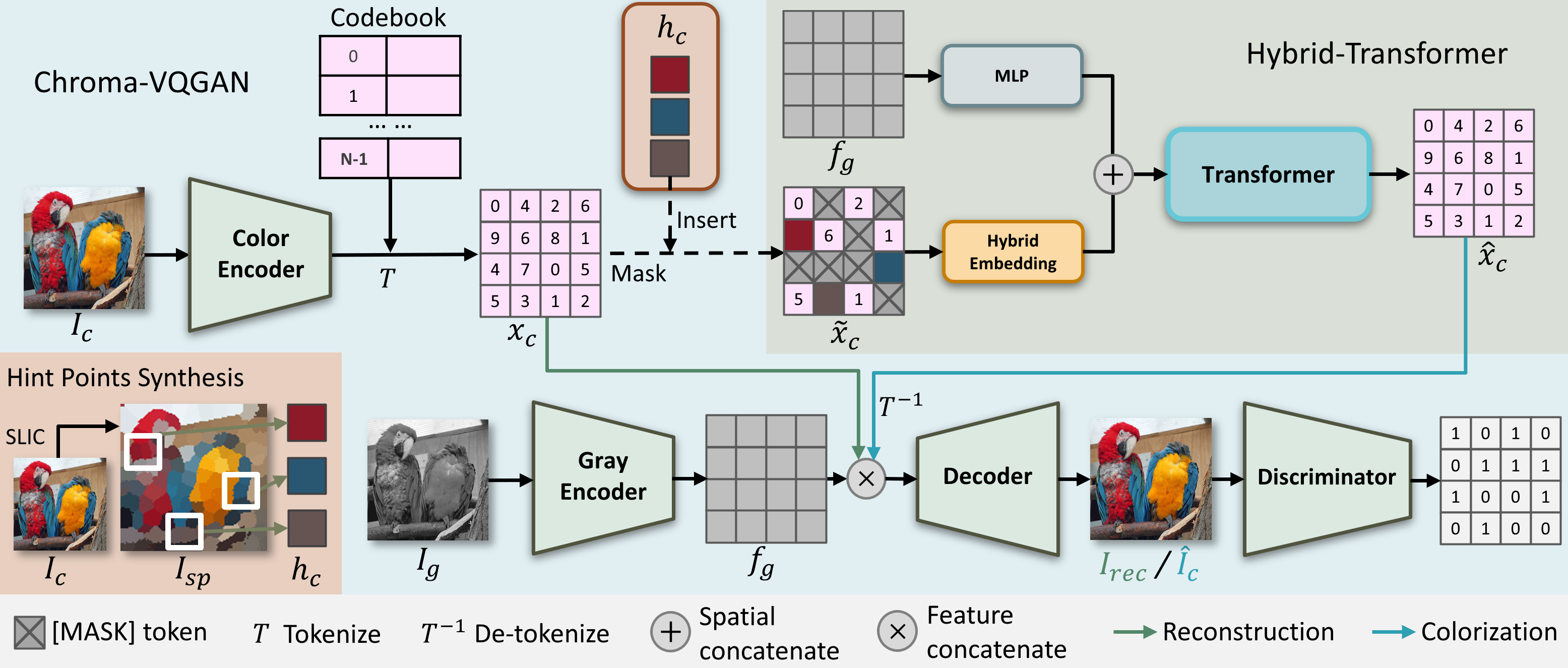}}
	\caption{Architecture of diverse unified colorization network. The network consists of two sub-nets: (a) a Chroma-VQGAN to disentangle and quantize chroma representation from the continuous gray one, and (b) a Hybrid-Transformer to generate diverse colorization results from unified conditions and continuous gray features. \rv{Input image: from ImageNet.} }
	\label{fig:structure}
\end{figure*}

Before introducing our Chroma-VQGAN, we first revisit the typical VQGAN~\cite{taming}. VQGAN encodes color images and quantizes them into a discrete codebook. Each image can be represented by a spatial collection of codebook entries. VQGAN learns the codebook through a reconstruction task by decoding such spatial collection with a decoder.
Since detailed structure information is lost during the quantization process, the reconstruction result can contain serious distortions and artifacts (will be validated in \refsec{abl_vqgan}). However, this is not acceptable in the colorization task, as the structure should be aligned well with the grayscale input.
Therefore, to adapt VQGAN to our task, we create a variant called Chroma-VQGAN for encoding gray and color features separately to preserve the detailed structure with continuous gray but quantized chrominance representations.

Given a color image $I_c$ with its gray version $I_g$ as inputs, Chroma-VQGAN takes two actions to mitigate the distortion artifacts and improve the reconstruction quality. First, we introduce a side branch, \ie~gray encoder shown in~\reffig{structure}, for extracting features from the gray input. Second, to avoid the information loss during quantization, we directly fuse the features of gray input without quantizing for sending it into the decoder to reconstruct the input color image. More formally, we first obtain color features $f_c$ from the input color image $I_c$ through the color encoder, while obtaining gray features $f_g$ from the input gray image $I_g$ through the gray encoder. To incorporate with the hint points, both features are down-sampled by the encoders with a factor of $d$, which is same as the cell size of the hint points. We then tokenize the color features $f_c$ into $x_c$ through a learnable codebook $\mathcal{Z}=\{z_k\}^{N-1}_{k=0}$, but remain the gray features as the continuous one. The tokenized $x_c \in \mathbb{R}^{H/d\times{W/d}}$ contains the indices of the entries from the learned codebook~\cite{taming} and the index of $i^{th}$ row and $j^{th}$ column is obtained by: 
\begin{equation}
    x_c^{ij} = T(f_c^{ij}) = \mathop{\arg\min}_{k \in [0, N-1]} \|f_c^{ij} - z_k\|,
\end{equation}
where $T(.)$ is a tokenization operation.

Before combing with the continuous gray features $f_g$, the color indices $x_c$ are detokenized by $T^{-1}$ for restoring the color features as:
\begin{equation}
    \hat{f}_c = T^{-1}(x_c) = \{ z_k \in \mathcal{Z}, k=x_c\}.
\end{equation}
The detokenized features $ \hat{f}_c$ are then concatenated with the gray features $f_g$ along the channel dimension. After decoding through a decoder, we obtain the reconstructed color image $I_{rec}$. The Chroma-VQGAN is trained in an adversarial way by learning accurate and perceptually rich features with the help of a discriminator, following the strategy in the previous work~\cite{taming}.

\rv{Unlike the previous work~\cite{taming} training separate VQGANs for quantizing both the condition and the image, our Chroma-VQGAN uses a single VQGAN with two encoder branches and the gray features are kept unquantized, as shown in \reffig{structure}. This design enables high-quality reconstructions, where the structure and content are well preserved from the input images.}
Another benefit of our Chroma-VQGAN is that it can disentangle the chrominance features in $f_c$ from the gray one $f_g$, because of the little information loss from gray image encoding (will be validated in \refsec{abl_vqgan}). This allows the Hybrid-Transformer introduced next to focus on the chrominance prediction, achieving better final results.

\subsubsection{BERT-Style Hybrid-Transformer.} \label{sec:s4_hybrid_transformer}

In this subsection, we introduce how we generate diverse colorization results from the unified hint points condition and grayscale input. We take a BERT-style scheme ~\cite{bert} for training our Transformer. During the training, the model needs to fulfill a color completion task. That is, we randomly mask out a portion of input color tokens, and ask the model to restore these tokens based on gray image, hint points, and the unmasked color tokens. Different from the traditional Transformer only relying on discrete color tokens as the inputs, our model has a hybrid input.

The hybrid inputs to our Transformer are extracted continuous gray features $f_g$, masked color tokens $\tilde{x}_c$, and hint points $h_c$, where $\tilde{x}_c=x_c^M \cup x_c^{\overline{M}}$, and $M$ indicates the indices of the tokens masked out. For each masked token, we replace the original codebook index with a special token $[MASK]$. Since our model should also support unconditional colorization without hint points during the inference time, we first introduce a simplified formulation, and then discuss how the hint points are synthesized with an adapted formulation. Conditioned on the unmasked color tokens $x_c^{\overline{M}}$, and the gray features $f_g$, our Hybrid-Transformer is trained to learn the likelihood of the indices at the masked positions for restoring $x_c$:
\begin{equation}  \label{eqn:bert_prob}
    P(x_c^M|x_c^{\overline{M}}, f_g) = \prod_{i \in M} P(x_c^{i}|x_c^{\overline{M}}, f_g).
\end{equation}
Then the Transformer is trained to minimize the softmax cross-entropy loss between the output probabilities and the ground-truth color indices.

\textbf{Hint Points Injection.} Rather than quantizing hint points as tokens to the Transformer, we directly use the continuous color values to keep the accurate color of the hint points. To better fuse the hint points $h_c$ with the color tokens $x_c$, we create a hybrid embedding space by mapping color tokens through an embedding layer for generating embedding color features $e_{x_c}$. We learn a mapping function from the color of hint points $h_c$ to the feature space $f_h$, sharing the same channel dimension $d_e$ as the embedded features $e_{x_c}$:
\begin{equation}
    f_h = W_h h_c + p_h,
\end{equation}
where $W_h \in \mathbb{R}^{3\times{d_e}}$ is the learnable weights and $p_h$ is a learnable positional embedding to distinguish the positions of hint points from color tokens. Note that we only insert the hint points at the masked positions $M$ during the training time. Given the mapped features of the hint points $f_h$ as another conditional input, the conditional probability of the Hybrid-Transformer in \refeqn{bert_prob} can be rewritten as:
\begin{equation}
    P(x_c^M|x_c^{\overline{M}},f_g,f_h) = \prod_{i \in M} P(x_c^{i}|x_c^{\overline{M}}, f_g, f_h).
\end{equation}

\textbf{Hint Points Synthesis.} Collecting a large scale annotated data covering all kinds of multi-modal conditions, and converting them into unified hint points are impractical, we thus create a novel method for synthesizing hint points during the training process. The key idea is to randomly sample grid cells from the original color image, and extract the hint point from each of these cells. For each selected cell, the color of the hint point should be sampled from the $d \times d$ image cell. A simple way is to take the mean color of the whole cell as the hint color, but we find this naive solution is problematic, as a single cell may cover multiple color regions and boundaries. As shown in \reffig{structure}, the cells of $I_{sp}$ in white frames may cover multiple colors, \eg~orange and blue, and the mean color cannot represent either one.

To avoid such vague and inaccurate color assignment, we propose to use the dominant color of each cell. We first compute the superpixel segmentation from the color image $I_c$ through Simple Linear Iterative Clustering (SLIC) Algorithm \cite{slic}. Then we get the superpixel image $I_{sp}$ by representing each superpixel with the mean color of the segment. During training, we randomly select a grid cell from the superpixel images, and take the dominant color value within that cell as the color of the hint point. To let the Hybrid-Transformer also deal with unconditional colorization in the inference time, there is a chance (\ie~30\%) that no hint point is added at the training time.

\subsubsection{Inference}

During the inference time, the model starts with the gray features $f_g$, hint points $h_c$ and color tokens $\tilde{x}_c$ all filled with $[MASK]$ tokens. Then the predicted color tokens $\hat{x}_c$ are sampled autoregressively (AR) in raster scan order with Hybrid-Transformer:
\begin{equation}
    p(\hat{x}_c|f_g,f_h) = \prod_{i} p(\hat{x}_c^{i}|\hat{x}_c^{<i}, f_g, f_h),
\end{equation}
where the probability of the current token is conditioned on all previously sampled tokens, the gray features, and the hint points. For each generated color token, we apply multinomial sampling within the top-k (\ie~k is set to 100) indices under the predicted probability distribution. We keep the hint point tokens fixed during the sampling process. After sampling all the tokens within $\hat{x}_c$, we detokenize the tokens to continuous features before concatenating them with the gray features along the channel dimension. Finally, the concatenated features are fed into the decoder of our Chroma-VQGAN to obtain the colorization image result $\hat{I}_c$.

\subsection{Interactive Interface}  \label{sec:interface}

\begin{figure}[t]
	\centerline{\includegraphics[width=1.0\linewidth]{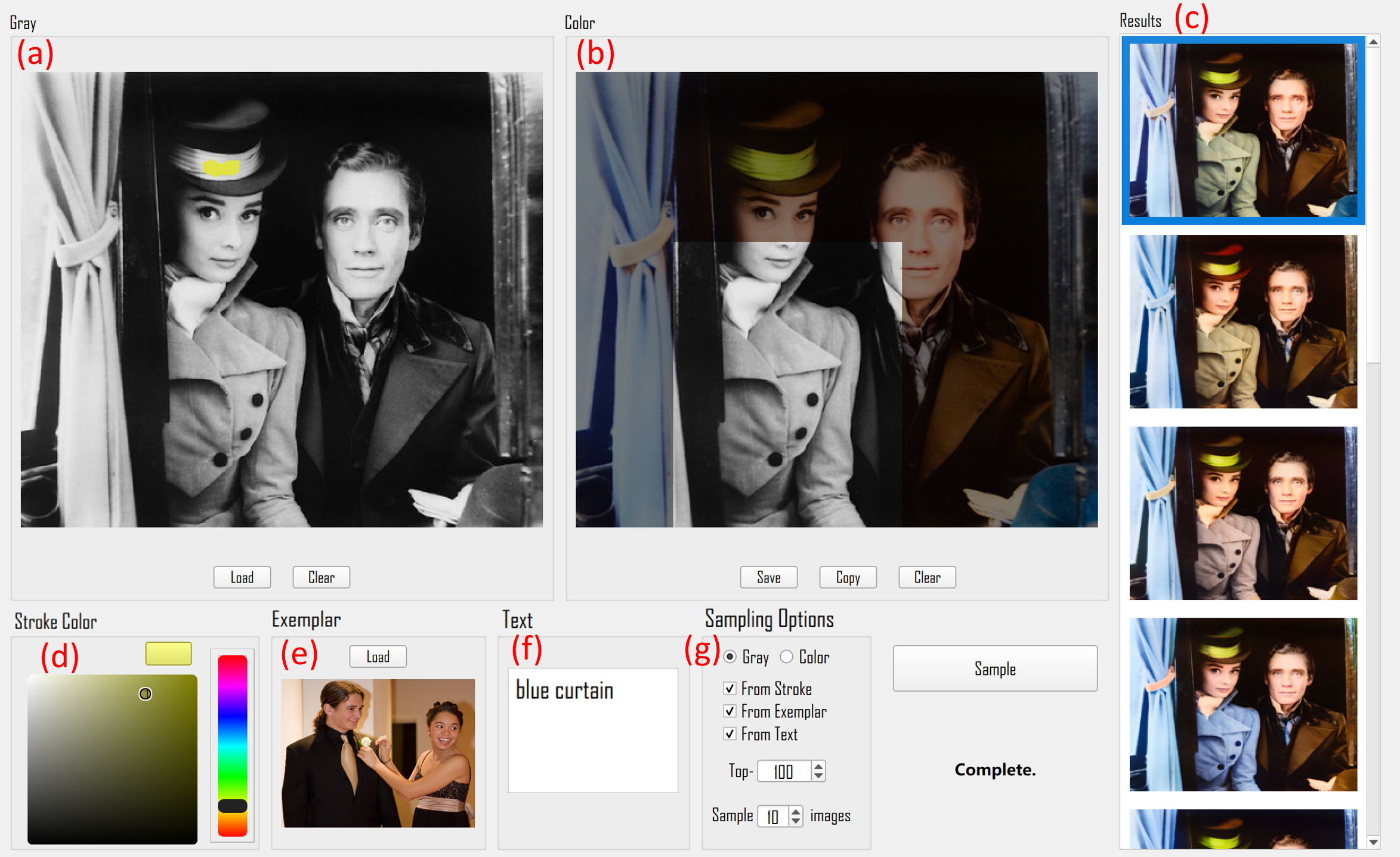}}
	\caption{Our interface of interactive multi-modal image colorization. (a) displays the input grayscale image; (c) shows all the diverse colorization results; (b) is the user-selected colorized image from (c), which can be further re-colorized and iteratively edited; (d) is the color picker of the stroke that can be drawn in (a) or (b) for stroke-based colorization; (e) is the input reference image for exemplar-based colorization; (f) is the input text description for text-based colorization; (g) is the panel for input and modality options. \rv{Input image: Audrey Hepburn and Mel Ferrer, while filming \textit{War and Peace}, 1955. Reference image: from MSCOCO.}}
	\label{fig:ui}
\end{figure}

Driven by our proposed UniColor framework, we design an interactive tool for multi-modal colorization. \reffig{ui} shows one screenshot of the user interface of our tool. Our tool supports various types of image colorization, including unconditional, stroke-based, exemplar-based, text-based, and hybrid colorization. In the hybrid colorization mode, the user can choose to add conditions in a combined way. Our tool has four main components: 1) a canvas for showing the grayscale input image and drawing strokes (a), 2) a panel showing all the diverse colorization results (c), 3) a canvas for re-colorizing the color image (b), and 4) an interface for inputting various types of conditions (d,e,f) and a panel for input and modality selection (g).

\textbf{Unconditional Colorization.} For unconditional colorization, the user just needs to input a gray image in (a) and uncheck all the conditions in (g) to let the system colorize the image automatically. The diverse results will be displayed in (c).

\textbf{Multi-modal Colorization.} Our system allows the user to select one modality for colorization from the panel (g) and then specify the conditions in different forms. For stroke-based colorization, the user can select colors in (d) and draw color strokes onto the grayscale image in (a); for exemplar-based colorization, the user can import the reference color image in (e); for the text-based colorization, the user can type in the text description of objects and colors in the textbox (f). Different types of conditions will be converted into hint points to guide the colorization process. 

\textbf{Hybrid Controls.} Our system also supports hybrid controls. The user can input more than one type of condition and check multiple modalities in (g), while the system will mix all the hint points generated from the selected conditions for colorization. To avoid conflict, we define the default priority of hybrid controls as: stroke, text, and exemplar, \eg~if stroke and exemplar conditions generate hint points on the same location, the points generated from the exemplar will be ignored, and only the points from the stroke will be considered.

\textbf{Re-colorizing \& Iterative Editing.} After colorization, the system will show the diverse colorization results in (c), and the user may select one to be displayed in (b). If the user wants to edit the selected result further, he or she could select an image subregion to re-colorize. Multi-modal conditions can also be applied to the subregion to reflect the user's intents. And the user may select and re-colorize subregions iteratively to interactively refine the result. Additionally, if a color image is imported, the system will also show the color version of the image in (b) so that the user could directly re-colorize an original color image in the same manner.

\section{Implementation Details}

\subsection{Network Architectures}
In this subsection, we introduce the detailed network architectures of both Chroma-VQGAN and Hybrid-Transformer.

\textbf{Chroma-VQGAN.} We mainly follow the implementations of the previous work~\cite{taming}. For the color encoder, we increase the number of channels from $256$ to $512$. For the quantization module, we set the size of the codebook as $N=4096$. For the gray encoder, we use the same structure as the color encoder, except changing the input channel to $1$. The input color images and grayscale images are down-sampled by a factor of $d=16$. After concatenating the gray and color features, the number of input channels to the decoder is $1024$.

\textbf{Hybrid-Transformer.} We use a similar structure as iGPT \cite{igpt} with learnable positional encoding. The Transformer consists of 24 multi-head self-attention (MHSA)~\cite{self-attention} layers with $16$ heads. The color tokens are embedded into features of channel dimension $d_e=512$. The gray features and color of hint points are passed through two separate linear layers, respectively, both with output dimensions of $512$. After that, the gray features $f_g \in \mathbb{R}^{{16}\times{16}\times{512}}$ and the embedded color features $e_{x_c} \in \mathbb{R}^{{16}\times{16}\times{512}}$ are flattened and concatenated in spatial dimension to form the input of shape $512\times{512}$ (\ie~$512=2\times 16^2$). We compute the attentions among all tokens without adding any mask. The final dimension of the predicted probability is the same as the length of the codebook (\ie~$4096$).

\subsection{Training Strategy}

\textbf{Training Dataset}. We train both the Chroma-VQGAN and Hybrid-Transformer on the ImageNet ILSVRC2012 \cite{ILSVRC} dataset, with around 12M training images. During training, the images are first resized to $294\times{294}$ and then randomly cropped to $256\times{256}$ followed by random horizontal flipping.

\textbf{Chroma-VQGAN.} The Chroma-VQGAN is trained on 4 Nvidia V100 GPUs with a total batch size of 32 for 260,000 steps (around 60 hours). The learning rate is set to $5.12\times{10^{-5}}$ (base learning rate $1.6\times{10^{-6}} \times$ batch size 32) throughout the whole training stage with no warm-up or decay. For stability, we begin to update the parameters of the discriminator after $6000$ steps.

\textbf{Hybrid-Transformer.} The hybrid-Transformer is trained in BERT-style \cite{bert}, where $16$ to $256$ randomly selected input color tokens are masked and replaced with a learnable $[MASK]$ token. We also apply an additional probability of $5\%$ to mask all the $256$ color tokens to ensure that the model can predict the colors from scratch. Among the masked tokens, we randomly select $1$ to $16$ positions to insert hint points, which are generated from the superpixel images as described in \refsec{s4_hybrid_transformer}. To ensure the transformer is capable of predicting color tokens without hint points, we only insert the hint points with a probability of $70\%$. All the numbers and the positions of the masking and the hint points are sampled from the uniform distribution. The hybrid-Transformer is trained on 4 Nvidia V100 GPUs with an accumulated batch size of $2\times{4}\times{16}=128$ for $142,000$ steps (around $46$ hours). The learning rate is set to $2.05\times{10^{-4}}$ (base learning rate $1.6\times{10^{-6}} \times$ batch size $128$) and decayed by 0.1 at $10^{th}$ epoch.

\rv{
\subsection{Inference \& Interaction Speed}
Our model takes an average of 4.6 seconds to colorize a $256\times{256}$ image with a single Nvidia V100 GPU, which is the common speed of an autoregressive Transformer. For text-based colorization, it will take an additional 1.8 seconds to convert the text to hint points. For the interactive system, it normally takes the user 10-15 seconds to input every single modality (\ie~stroke, exemplar, and text).  
}

\section{Experiments}

In this section, we first compare our unified framework with previous works in all four types of colorization (unconditional, stroke-based, exemplar-based, and text-based) respectively (\refsec{comparison}). We then conduct the user study and show the results in \refsec{user_study}. Lastly, we perform ablation studies in \refsec{ablation}.

\begin{table}[t]
\small
\caption{Comparison with unconditional colorization methods on both ImageNet and MSCOCO datasets. The models are trained on ImageNet if not specified (* Trained on MSCOCO). Metrics calculated on original images are taken as references.}
\label{tab:uncond_metric}
\begin{minipage}{\columnwidth}
\begin{center}
\begin{tabular}{ccccc}
  \toprule
    {} & \multicolumn{2}{c}{ImageNet} & \multicolumn{2}{c}{MSCOCO} \\
    {} & FID$\downarrow$ & Colorful$\uparrow$ & FID$\downarrow$ & Colorful$\uparrow$ \\ \toprule
    Original 
    & --- & 38.00 & --- & 37.46 \\ \midrule \midrule
    CIC~\shortcite{colorful} 
    & 21.31 & 34.25 & 32.62 & 34.36 \\ \midrule
    User-guided (auto)~\shortcite{user-guided} 
    & 12.53 & 26.16 & 19.18 & 27.04 \\ \midrule
    Deoldify~\shortcite{deoldify} 
    & 9.59 & 21.39 & 12.29 & 22.84 \\ \midrule
    InstColor*~\shortcite{instance-colorization} 
    & 12.74 & 26.00 & 12.72 & 27.26 \\ \midrule
    ChromaGAN~\shortcite{chromagan} 
    & 16.27 & 26.92 & 25.50 & 27.08 \\ \midrule
    GenPrior~\shortcite{vivid-colorization} 
    & 9.57 & 35.29 & --- & --- \\ \midrule \midrule
    Coltran~\shortcite{coltran} 
    & 12.31 & 36.59 & 14.20 & 36.31 \\ \midrule \midrule
    Ours 
    & \textbf{9.46} & \textbf{39.01} & \textbf{11.16} & \textbf{39.11} \\
  \bottomrule
  \end{tabular}
\end{center}
\end{minipage}
\end{table}

\textbf{Test Data.} We select testing images from two datasets: ImageNet ILSVRC2012~\cite{ILSVRC} and MSCOCO ~2017~\cite{coco}. For ImageNet, we randomly select $5,000$ images from the $50,000$ images in the validation set, where 5 images are drawn from each class. For MSCOCO, we use all the $5,000$ images from the validation set with ground-truth text caption and segmentation.

\subsection{Comparisons on Multi-modal Colorization}
\label{sec:comparison}

As a unified multi-modal colorization framework, we need to examine the performance on different condition modalities, and we first compare with previous state-of-the-art (SOTA) methods on: unconditional, stroke-based, exemplar-based, and text-based colorization, respectively. We show both quantitative and qualitative results for each of the conditions.

\textbf{Evaluation Metrics.} To measure the overall quality and fidelity of the generated images, we calculate \textbf{FID}~\cite{fid} between the generated and ground-truth images. We also calculate \textbf{Colorfulness}~\cite{colorfulness} to measure how colorful and vivid are the colorized images. For stroke-based colorization, we further calculate \textbf{LPIPS}~\cite{lpips} to measure the perceptual similarity between the colorized and original images, because the generated images are supposed to be close to the original one, from which the hint points are sampled. For exemplar-based colorization, we introduce \textbf{Contextual Loss}~\cite{contextual} with pixel-wise L2-loss to measure the similarity between the non-aligned images, \ie~the colorized images and reference images. For text-based colorization, we use \textbf{CLIP score}~\cite{clip} to measure the relevance between the colorized images and the text prompts. All the metrics are calculated with images of size $256\times{256}$, except $96\times96$ for Contextual Loss.
\begin{figure*}[!t]
	\centerline{\includegraphics[width=1.0\textwidth]{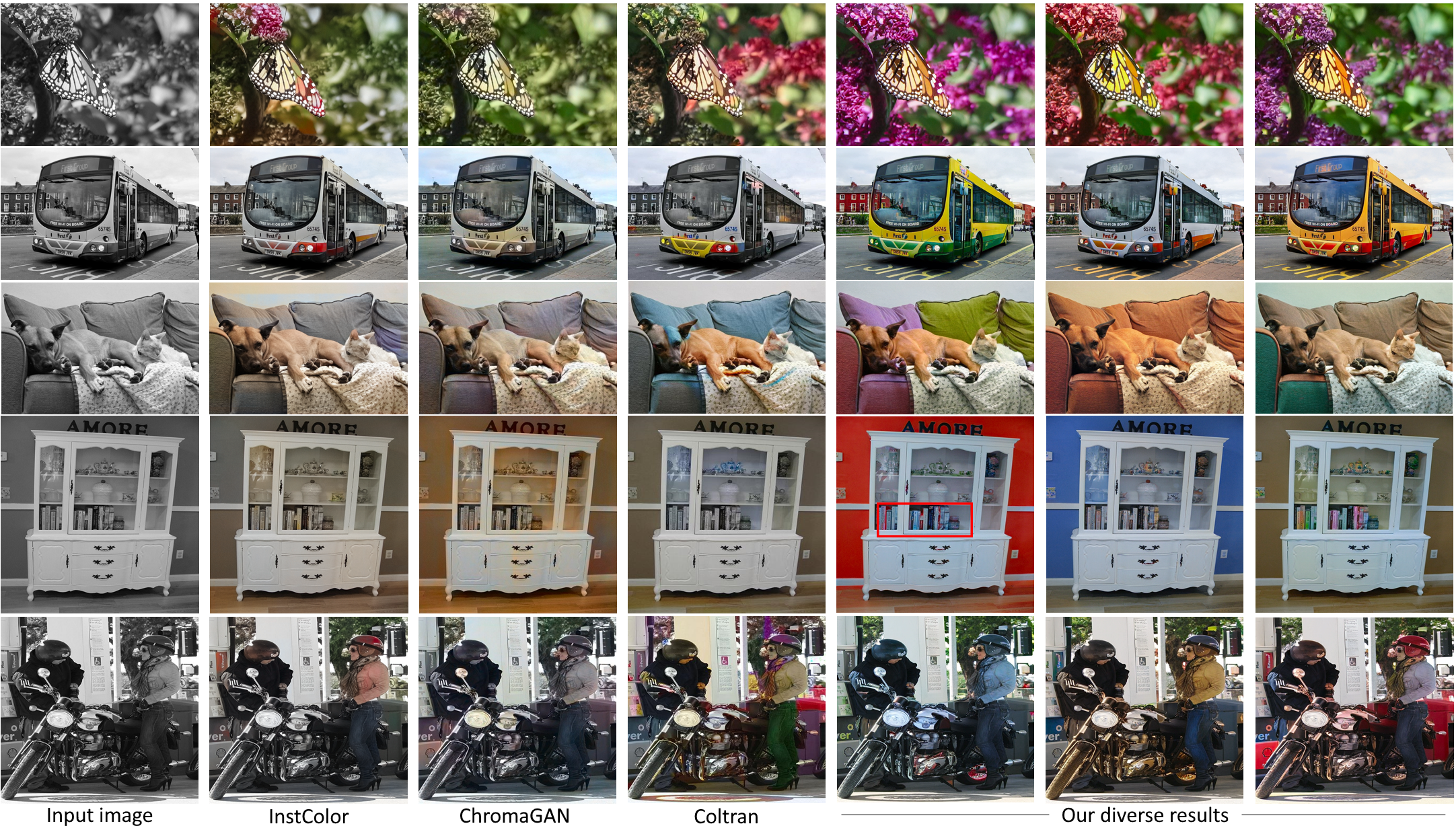}}
	\caption{Comparison with unconditional colorization methods: InstColor~\cite{instance-colorization}, ChromaGAN~\cite{chromagan}, and Coltran~\cite{coltran}. Our model can generate diverse results for each of the input grayscale images. \rv{Input images: the $1^{st}$ and $4^{th}$ rows are from ImageNet; others are from MSCOCO.} }
	\label{fig:uncond_comp}
\end{figure*}

\textbf{Unconditional Colorization.} \rv{For unconditional colorization, we compare with three types of SOTA methods} that can colorize grayscale images without user hints: a) CNN-based methods: CIC \cite{colorful}, User-guided (auto)~\cite{user-guided}, InstColor~\cite{instance-colorization}, Deoldify (software)~\cite{deoldify}, ChromaGAN~\cite{chromagan} and GenPrior~\cite{vivid-colorization}; b) Transformer-based method: Coltran~\cite{coltran}; \rv{and c) Diffusion-based model: Palette \cite{palette}}.

Quantitatively, as shown in \reftab{uncond_metric}, our method outperforms CNN-based and Transformer-based methods on both datasets in terms of FID and colorfulness. Generally, Transformer-based methods (Coltran and ours) with multinomial sampling produce more vivid and colorful colors than CNN-based methods, except that CIC encourages rare colors in the loss function (also observed in~\cite{exemplar-video, vivid-colorization}) and GenPrior uses auto-generated reference images. We show the qualitative comparison in \reffig{uncond_comp}.~\footnote{Because of the limited space, we only choose two representatives from CNN-based methods for qualitative comparison and leave the full results in the supplementary document.} Our method can generate diverse and vivid colors, with multinomial sampling from Hybrid-Transformer (\eg~diverse colors of the bus in $2^{nd}$ row and the pillows in $3^{rd}$ row). Thanks to the global attention module, different from CNN-based methods, our Transformer-based framework generates consistent color across distant pixels sharing the same semantics (\eg~the flowers in $1^{st}$ row and the sofa in $3^{rd}$ row). Compared with Coltran, the existence of CNN-based Chroma-VQGAN makes our model more sensitive to local contours (\eg~the dog and sofa in $3^{rd}$ row) and details (\eg~the books in $4^{th}$ row).

\begin{figure}[t]
	\centerline{\includegraphics[width=1.0\linewidth]{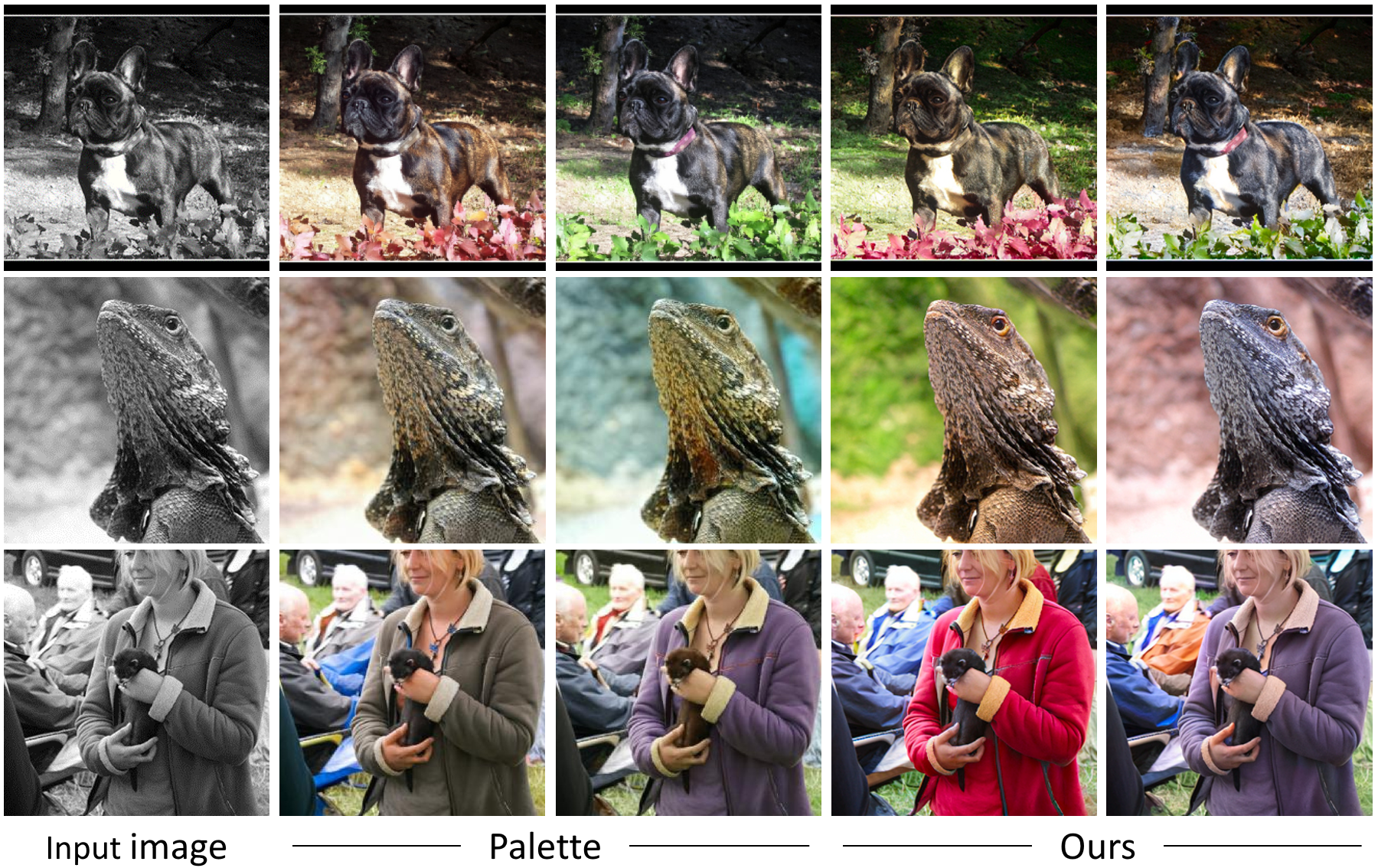}}
	\caption{\rv{Comparison with diffusion-based model Palette~\cite{palette}. Our method generates diverse results comparable to Palette. \rv{Input images: from ImageNet.} }}
	\label{fig:palette_comp}
\end{figure}

\rv{As the code of Palette~\cite{palette} is not available, we only compute the FID score following the evaluation protocol used in both Coltran and Palette on ImageNet validation set. We obtain the FID scores as: 19.37 (Coltran), 15.78 (Palette), and 16.80 (Ours). Under this protocol, our unconditional method outperforms Coltran and is comparable to Palette. For qualitative comparison, we demonstrate the results on the images shown in the original Palette paper. As can be seen in \reffig{palette_comp}, both Palette and our method can produce diverse colorization results with high fidelity and vivid colors. Besides, our method can further support multi-modal control.}

\textbf{Stroke-based Colorization.} For stroke-based colorization, we compare with User-guided~\cite{user-guided}, a recent SOTA work. To be fair, instead of generating the hint points from the superpixel images as specified in \refsec{s4_hybrid_transformer}, we adopt the same method as the User-guided one ~\cite{user-guided} to assign the mean color of a cell to the hint point. To avoid selecting hint points on the intersections of two objects, we first perform clustering and segmentation on the images with mean-shift and then randomly select $2$ to $16$ cells from the large segments as the hint points.

\begin{figure}[t]
    \centering
     \includegraphics[width=1.0\linewidth]{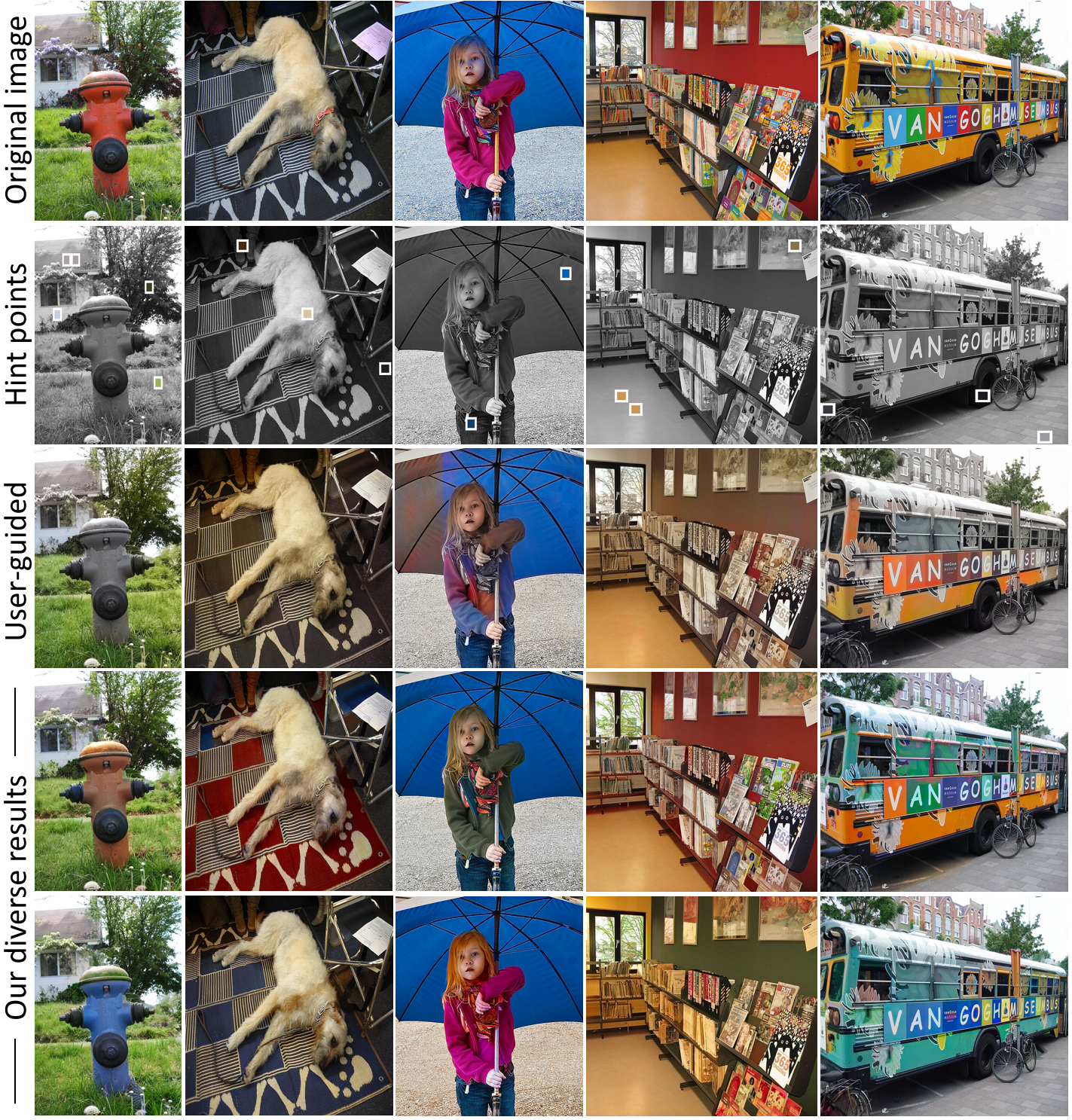}
     \caption{Comparison with the stroke-based method (User-guided~\cite{user-guided}). \rv{Input images: the $2^{nd}$ and $4^{th}$ columns are from ImageNet; others are from MSCOCO.}}
    \label{fig:stroke_comp}
\end{figure}

We show the quantitative results in \reftab{stroke_metric}. To verify whether our model follows the input condition, we also compare with our unconditional variant. As can be seen, the lower FID and LPIPS indicate that our model propagates the hint points derived from strokes properly. Compared with the previous stroke-based method~\cite{user-guided}, ours achieves better FID and colorfulness, and comparable LPIPS. As shown in \reffig{stroke_comp}, our method propagates the stroke color (hint points) into the whole object smoothly and consistently (\eg~the blue umbrella in $3^{rd}$ column), while User-guided fails to spread the hint colors to the whole object. For the region without specified hint points, our method can also generate diverse and vivid colors (\eg, the books and wall in $4^{th}$ column and the bus in $5^{th}$ column).

\begin{table}[t]
  \setlength\tabcolsep{3.1pt} 
  \small
  \caption{Comparison with the stroke-based colorization method - User-guided~\cite{user-guided} on both ImageNet and MSCOCO dataset. The unconditional variant (\ie~Uncond.) helps on verifying the effectiveness of input conditions.}
  \label{tab:stroke_metric}
  \begin{minipage}{\columnwidth}
  \begin{center}
  \begin{tabular}{cccccccc}
      \toprule
      {} & \multicolumn{3}{c}{ImageNet} & {} & \multicolumn{3}{c}{MSCOCO} \\
      {} & FID$\downarrow$ & LPIPS$\downarrow$ & Colorful$\uparrow$& {} & FID$\downarrow$ & LPIPS$\downarrow$ & Colorful$\uparrow$ \\ 
      \toprule
      User-guided & 9.76 & 0.1144 & 32.48 & {} & 14.68 & \textbf{0.1166} & 31.44 \\ 
      \midrule
      Ours (Uncond.) & 9.46 & 0.1945 & \textbf{39.01} & {} & 11.16 & 0.1909 & \textbf{39.11} \\
      \midrule
      Ours (Stk.) & \textbf{7.04} & \textbf{0.1119} & 36.16 & {} & \textbf{8.89} & 0.1189 & 35.71  \\ 
      \bottomrule
  \end{tabular}
  \end{center}
  \end{minipage}
  \end{table}

  \begin{table}[t]
  \small
  \caption{Comparison with the exemplar-based methods on ImageNet.}
  \label{tab:exp_metric}
  \begin{minipage}{\columnwidth}
  \begin{center}
  \begin{tabular}{cccc}
    \toprule
      {} & FID$\downarrow$ & Contextual$\downarrow$ & Colorful$\uparrow$ \\ \toprule
      Deep Exp.~\shortcite{exemplar-image} & 10.79 & \textbf{1.75} & 29.64 \\ \midrule
      Exp. Video~\shortcite{exemplar-video} & 10.70 & 1.90 & 25.92 \\ \midrule
      Ours (Uncond.) & 9.46 & 2.65 & \textbf{39.01} \\ \midrule
      Ours (Exp.) & \textbf{7.39} & 1.82 & 38.80 \\
    \bottomrule
  \end{tabular}
  \end{center}
  \end{minipage}
  \end{table}

  \begin{table}[t]
  \small
  \caption{Quantitative results of text-based colorization on MSCOCO.}
  \label{tab:text_metric}
  \begin{minipage}{\columnwidth}
  \begin{center}
  \begin{tabular}{ccc}
    \toprule
      {} & CLIP similarity$\uparrow$ & FID$\downarrow$ \\ \toprule
      Original & 24.05 & --- \\ \midrule
      Ours (Uncond.) & 23.55 & \textbf{11.16} \\ \midrule
      Ours (Text) & \textbf{24.50} & 11.29 \\
    \bottomrule
  \end{tabular}
  \end{center}
  \end{minipage}
  \end{table}

\textbf{Exemplar-based Colorization.} For exemplar-based colorization, we compare with Deep Exp. \cite{exemplar-image} and Exp. Video \cite{exemplar-video}. We only test on ImageNet because similar reference images could hardly be obtained for images in MSCOCO. We use the retrieval method based on gray images in \cite{exemplar-image} to obtain the reference images from the ImageNet training set, which are closest to the input grayscale images.

\begin{figure}[t]
    \centering
     \includegraphics[width=1.0\linewidth]{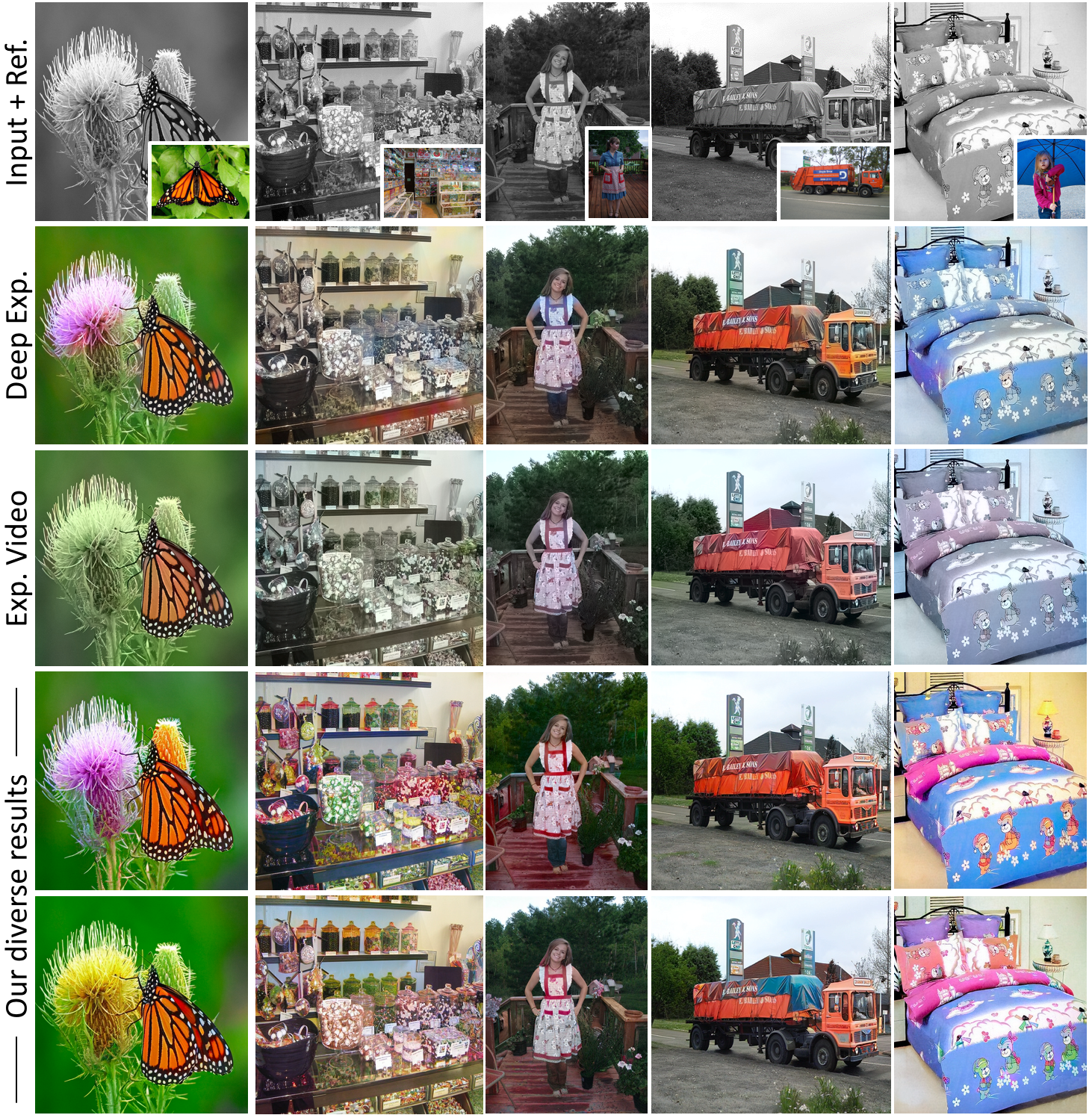}
     \caption{Comparison with exemplar-based methods (Deep Exp.~\cite{exemplar-image} and Exp. Video~\cite{exemplar-video}). The reference images are shown in the right-bottom corner of each input grayscale image. \rv{Input images: the reference image in the $5^{th}$ column is from MSCOCO, and the other images are from ImageNet.} }
 	\label{fig:exemplar_comp}
\end{figure}

As shown in \reftab{exp_metric}, our exemplar-based method obtains lower FID and contextual loss than our unconditional variant, which demonstrates the effectiveness of the reference images. Compared with other exemplar-based methods, ours achieves the best FID and colorfulness scores. In the aspect of contextual loss, ours performs a bit worse than Deep Exp. \cite{exemplar-image}, due to the fact that our method selectively inherits the colors from the warped image with high confidence, as stated in \refsec{hint_point}. This mechanism makes our method more robust when the warped image is unreliable. As shown in $3^{rd}$ column of \reffig{exemplar_comp}, the color of the girl's arm is not misguided by the wrongly warped blue. When the warped image of the sugar store in $2^{nd}$ column and the bed in $5^{th}$ column are noisy, other methods generate results of low colorfulness, while our method still generates images with vivid and diverse colors.


\begin{figure}[!t]
	\centerline{\includegraphics[width=1.0\linewidth]{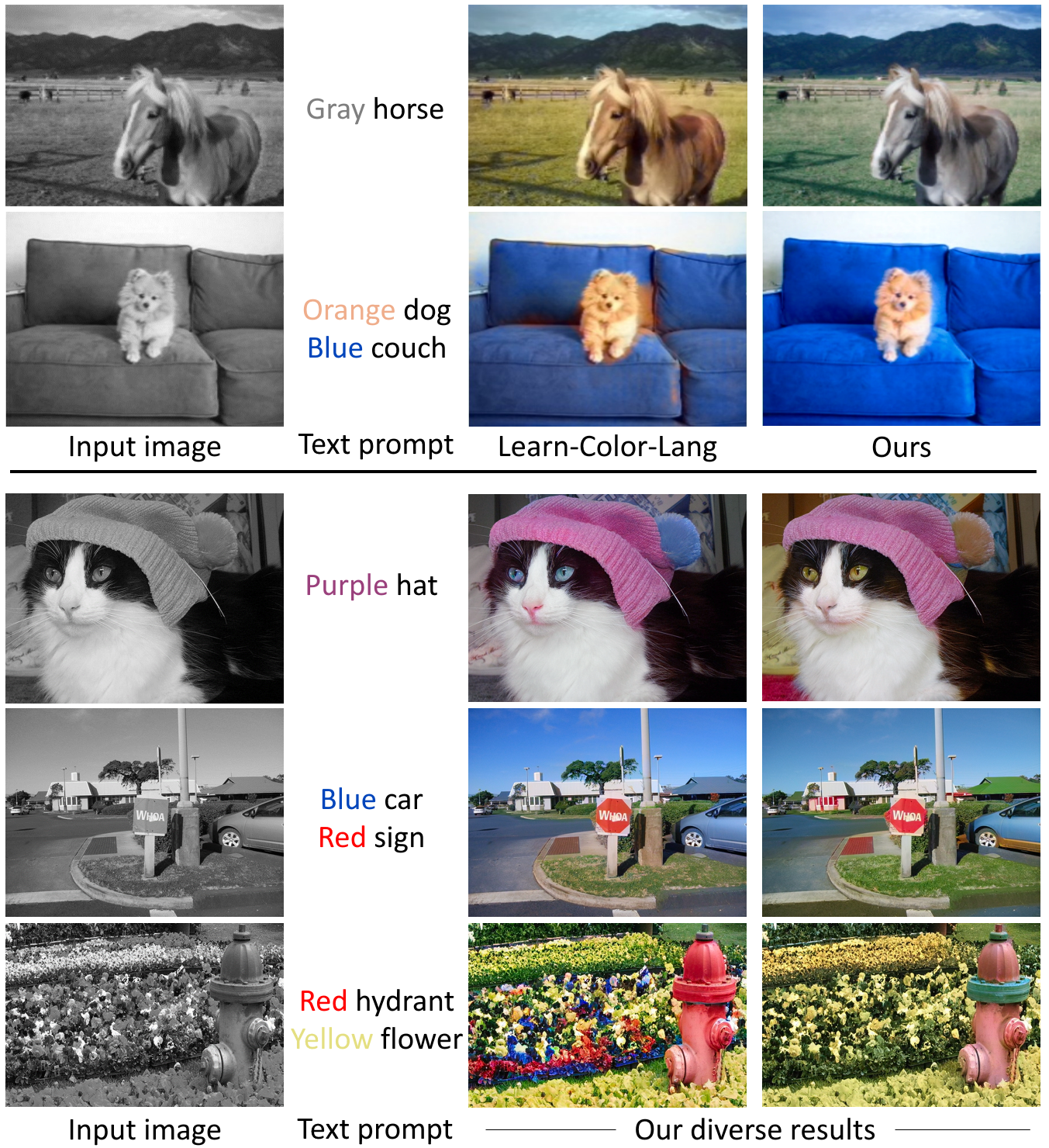}}
	\caption{Comparison with the text-based colorization method Learn-Color-Lang~\cite{learn-color-language}. We show the comparison in the first group, and our diverse results in the second group. \rv{Input images: $1^{st}$ group is from paper~\cite{learn-color-language}, $2^{nd}$ group is from MSCOCO.} }
	\label{fig:text_comp}
\end{figure}

\textbf{Text-based Colorization.} For text-based colorization, we only test MSCOCO with ground-truth text caption and segmentation. Because most of the text captions include no color descriptions, we automatically insert a color word before each object word. We first cluster the RGB values of all the pixels within the ground-truth segmentation of each object, and then assign the color with the highest occurrence to that object. Then we colorize the images from the new captions with color descriptions. Since the implementation and full results of the previous text-based method Learn-Color-Lang \cite{learn-color-language} are not publicly available, we only perform visual comparison on the images shown in their original paper in \reffig{text_comp}.

\rv{Thus, for quantitative results, we only compare with our unconditional variant to verify whether the model follows the input text, and the score of ground-truth images is shown for reference. As shown in \reftab{text_metric}, our text-based method gets a higher CLIP similarity score and similar FID compared with the unconditional method. This indicates that our text-based method can correctly locate the objects and colorize them with accurate colors. Because of the ambiguities of color words, though our method generates different colors from the ground truths, they are still aligned well with the input color words. This may be the reason for the higher CLIP similarity score compared to the ground truths.}
Compared with Learn-Color-Lang (\reffig{text_comp}), our method produces more accurate color (\eg~the gray horse in $1^{st}$ row), with less artifacts (\eg~the blue sofa in $2^{nd}$ row). With CLIP-based hint points conversion, our method can respond to tiny objects (\eg~the red sign in $4^{th}$ row) and open-vocabulary objects which may not appear in MSCOCO (\eg~the purple hat in $3^{rd}$ row). This reflects the feasibility and flexibility of our method.

\subsection{User Study} \label{sec:user_study}

To further validate the effectiveness of our method, we conduct a user study and discuss the results in this subsection. As text-based colorization has no suitable previous work for comparison, we mainly compare three different design scenarios: unconditional, stroke-based, and exemplar-based colorization through the user study. For unconditional colorization, we select the two representative works used in qualitative comparison from CNN-based methods: InstColor~\shortcite{instance-colorization} and ChromaGAN~\shortcite{chromagan}, and Coltran~\shortcite{coltran} for Transformer-based method.

For each modality, we randomly collected $30$ design cases based on different input grayscale images. For unconditional and stroke-based ones, we selected 15 images per dataset from ImageNet and MSCOCO, while for exemplar-based methods, we selected all from the MSCOCO dataset. For each design case, given the input grayscale image, hint points for stroke-based colorization, and a reference image for exemplar-based colorization, we showed the results from different methods in random order. The participants were asked to select the best one based on two metrics: realistic and consistent with input condition (if it has). Each participant was asked for 8 formal questions per scenario with an additional validation question, deriving 25 questions in total. The validation question is very simple by putting the only ground truth image among grayscale ones for testing the faithfulness of each participant.

\begin{figure}[t]
	\centerline{\includegraphics[width=1.0\linewidth]{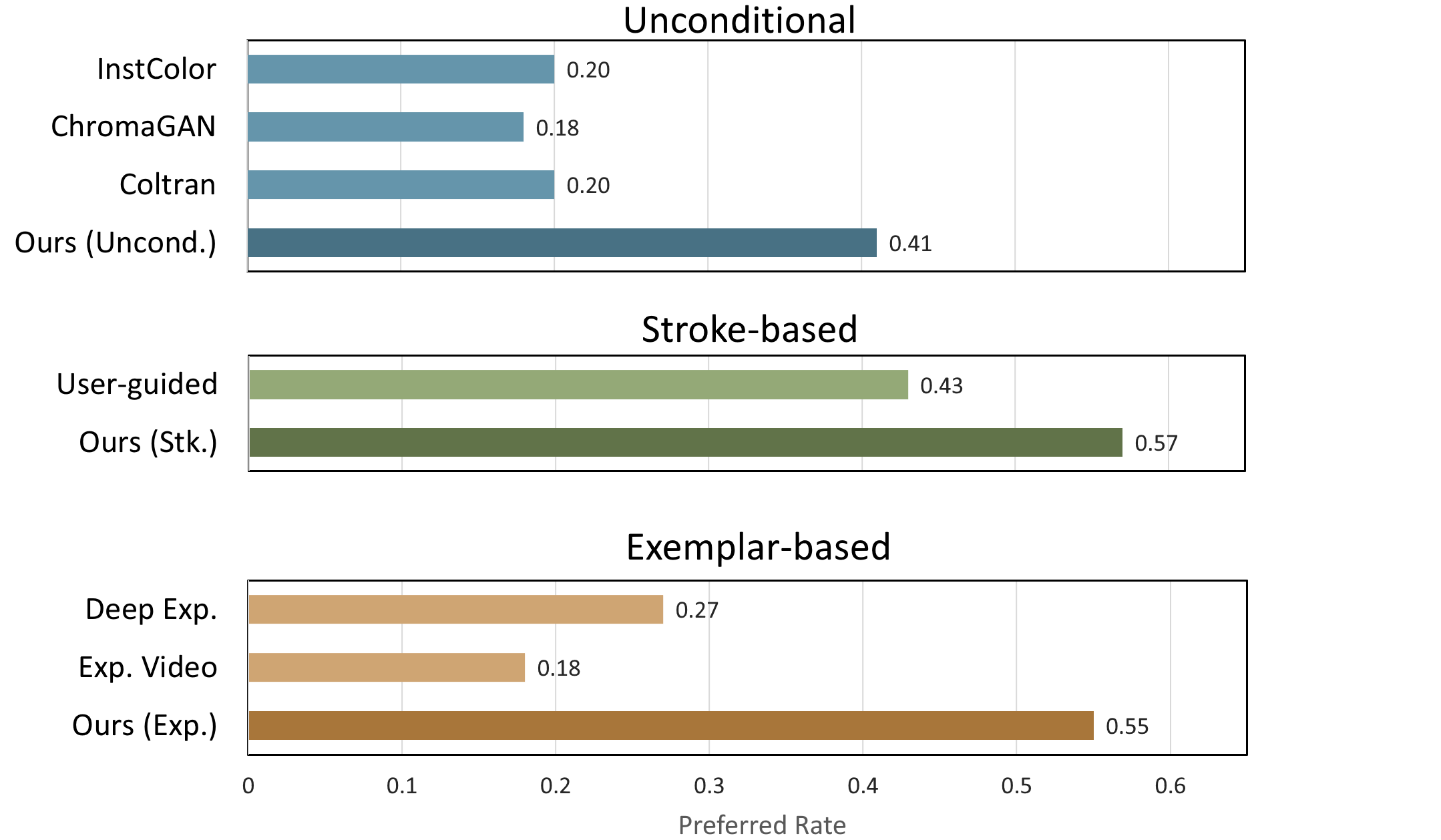}}
	\caption{Results of the user study on unconditional, stroke-based, and exemplar-based colorization tasks.}
	\label{fig:user_study}
\end{figure}

We finally collected $390$ questionnaires, where $301$ of them are valid with passing the validation question. Among the 301 participants, $163$ participants are below $20$ years old, $70$ range from $20$ to $40$ years of age, and $68$ are above 40 years old. We show the result of the user study in \reffig{user_study}, and we find that our method outperforms other methods in all unconditional, stroke-based, and exemplar-based colorization with a preferred rate of 41\%, 57\%, and 55\%, respectively. 

\subsection{Ablation Study} \label{sec:ablation}

\subsubsection{The effect of Chroma-VQGAN} \label{sec:abl_vqgan}

\begin{figure}[t]
	\centerline{\includegraphics[width=1.0\linewidth]{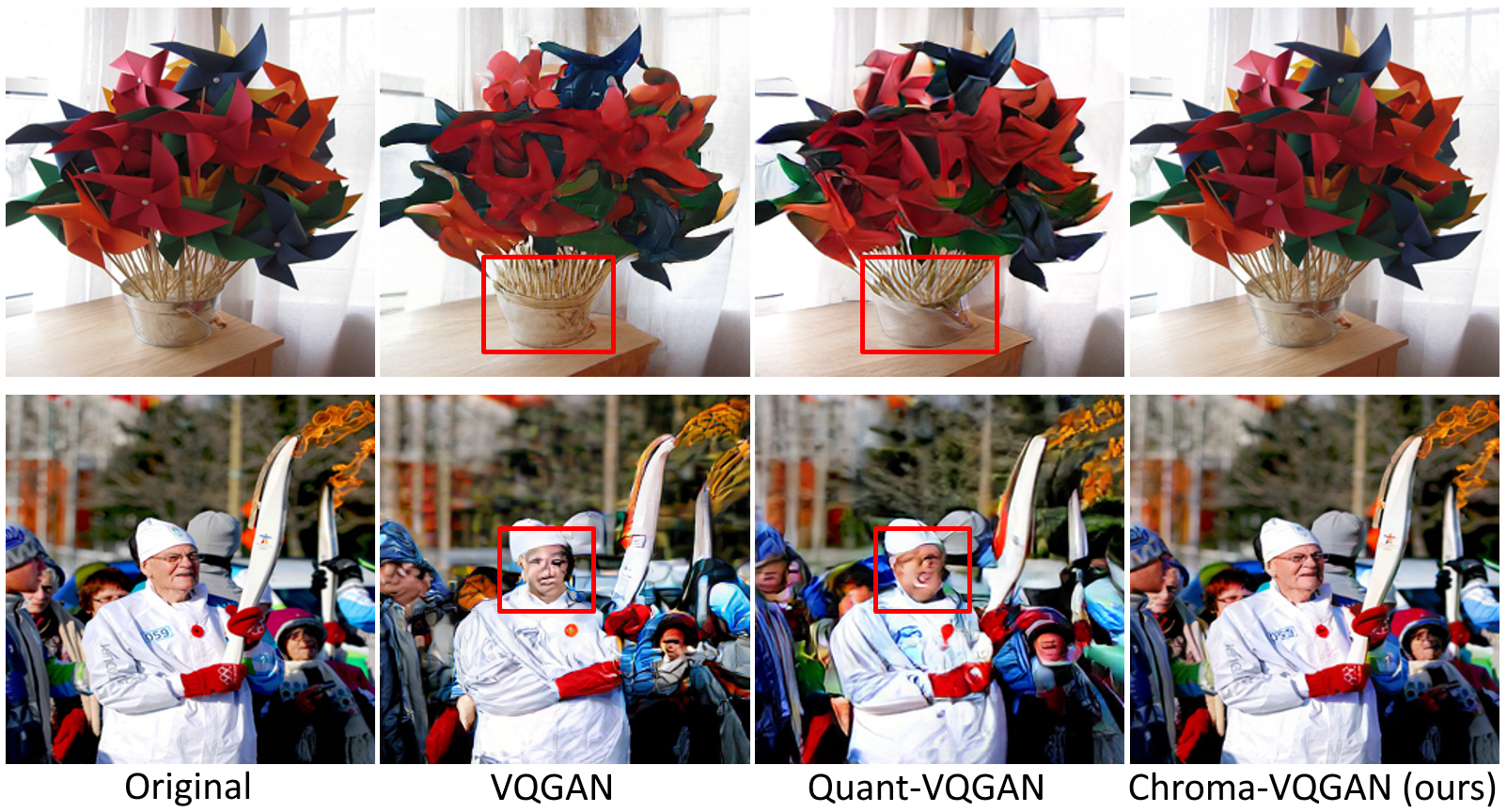}}
	\caption{Comparison results on image reconstruction for showing the effect of Chroma-VQGAN. \rv{Input original images: from ImageNet.} }
	\label{fig:vqgan_comp}
\end{figure}

To verify whether our Chroma-VQGAN can better reconstruct the images with unquantized gray features, we train a vanilla VQGAN and Quant-VQGAN, which has the same architecture as Chroma-VQGAN, but the gray features are quantized into tokens same as the color tokens. To compare the performance, we reconstruct the $5,000$ images from ImageNet and compute the FID, PSNR, LPIPS, and SSIM~\cite{ssim}. As shown in \reftab{ablation_recon} and \reffig{vqgan_comp}, our Chroma-VQGAN reconstructs the color image with less distortions and obtains better metrics, compared with both the vanilla VQGAN and Quant-VQGAN. \rv{Our method is even better than vanilla VQGAN with a 4 times smaller downsampling rate and 16 times larger number of tokens (leads to 256 times more computational cost and 4096 times more inference time). The experiment indicates that the additional unquantized gray features preserve the structural details during the decoding of the color tokens, which enables better reconstruction with much less computational cost.}

\begin{figure}[t]
	\centerline{\includegraphics[width=1.0\linewidth]{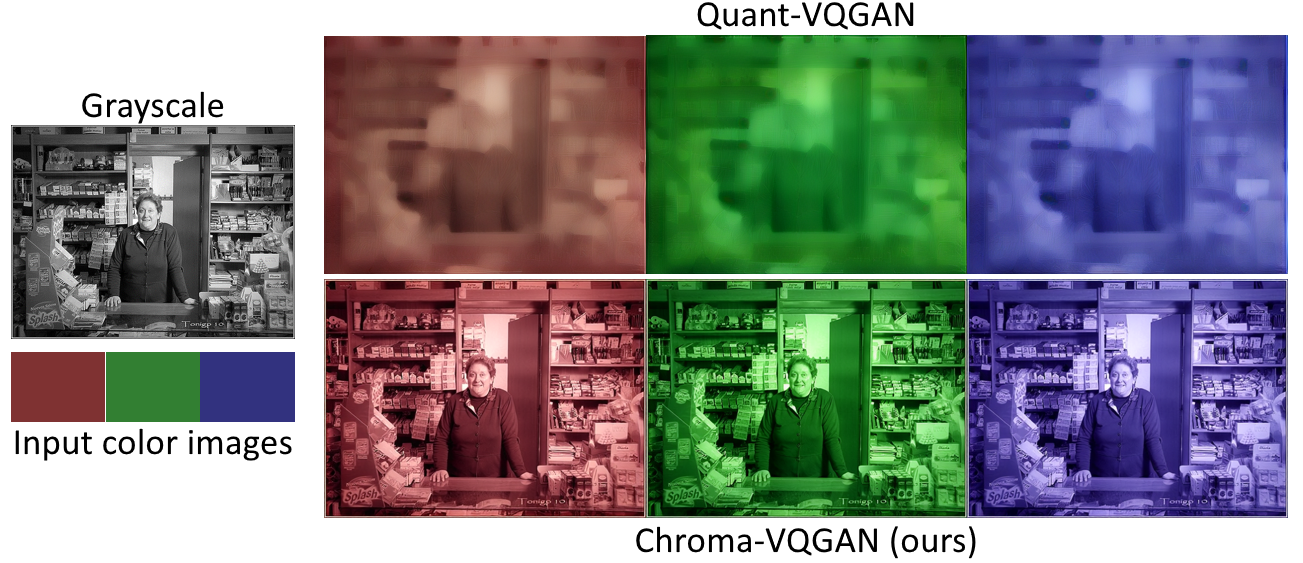}}
	\caption{Results of changing the input color images. Our Chroma-VQGAN generates colored images with high quality while results from Quant-VQGAN are blurred. \rv{Input image: from ImageNet.} }
	\label{fig:rgb_recon}
\end{figure}

To examine whether our Chroma-VQGAN learns disentangled chrominance representation in color tokens, we combine different pure color images with the input grayscale image. A good disentangled chrominance representation only controls the color without influencing the underlying structure. As shown in \reffig{rgb_recon}, our Chroma-VQGAN still preserves the structure details even though the input color image is changed, whereas the results of Quant-VQGAN are blurred. This implies that the color tokens in Chroma-VQGAN contain solely the chrominance information, whereas the color tokens in Quant-VQGAN still carry structure details. Therefore, by keeping gray features unquantized, the disentangled chrominance representation can help Hybrid-Transformer focus on color prediction without distracting from structural details.

\begin{table}[!t]
\small
\caption{Comparison results on image reconstruction. We compare our Chroma-VQGAN with the vanilla VQGAN~\cite{taming} and a variant called Quant-VQGAN.}
\label{tab:ablation_recon}
\begin{center}
\resizebox{\columnwidth}{!}{
\begin{tabular}{cccccc}
    \toprule
    {} & \makecell{Downsample \\ rate} & FID$\downarrow$ & PSNR$\uparrow$ & LPIPS$\downarrow$ & SSIM$\uparrow$ \\ \toprule
     \rv{VQGAN} 
     & \rv{ \begin{tabular}{@{}c@{}} 4$\times$ \\ 16$\times$ \end{tabular} } 
     & \rv{ \begin{tabular}{@{}c@{}} 2.61 \\ 11.83 \end{tabular} } 
     & \rv{ \begin{tabular}{@{}c@{}} 28.70 \\ 20.03 \end{tabular} } 
     & \rv{ \begin{tabular}{@{}c@{}} 0.0433 \\ 0.1691 \end{tabular} } 
     & \rv{ \begin{tabular}{@{}c@{}} 0.8555 \\ 0.5062 \end{tabular} } 
     \\ \midrule
    Quant-VQGAN & 16$\times$ & 11.78 & 20.67 & 0.1592 & 0.5316 \\ \midrule
    Chroma-VQGAN (Ours) & 16$\times$ & \textbf{1.68} & \textbf{29.73} & \textbf{0.0304} & \textbf{0.8770} \\
    \bottomrule
\end{tabular}
}
\end{center}
\end{table}

\begin{table}[t]
\small
\caption{The effect of Hybrid-Transformer. We compare with variants of replacing continuous gray features with discrete tokens (Quant-gray) and replacing continuous hint points with discrete tokens (Quant-hint).}
\label{tab:ablation_tran}
\begin{minipage}{\columnwidth}
\begin{center}
\begin{tabular}{cccc}
  \toprule
     & FID$\downarrow$ & LPIPS$\downarrow$ & Colorful$\uparrow$ \\ \toprule
    Quant-gray (Uncond.) & 11.88 & 0.2035 & \textbf{39.33} \\ \midrule
    Ours (Uncond.) & \textbf{9.46} & \textbf{0.1945} & 39.01 \\ \midrule \midrule
    Quant-hint (Stk.) & 9.76 & 0.1445 & 24.45 \\ \midrule
    Ours (Stk.) & \textbf{7.04} & \textbf{0.1119} & \textbf{36.16} \\
  \bottomrule
\end{tabular}
\end{center}
\end{minipage}
\end{table}

\subsubsection{The effect of Hybrid-Transformer}
Our Hybrid-Transformer takes inputs in a hybrid format with quantized color tokens, continuous gray features, and color hint points. In this subsection, we first create a baseline by replacing continuous gray features with discrete tokens based on Quant-VQGAN (\textbf{Quant-gray}) and test under unconditional colorization (\ie~without input color hints). We then create the other baseline (\textbf{Quant-hint}) by replacing the continuous color hints with color tokens. Specifically, we obtain the token index by sending the pure hint color image into the color encoder of Chroma-VQGAN. Note that the gray features remain continuous under hint points in discrete tokens.

\begin{figure}[t]
	\centerline{\includegraphics[width=1.0\linewidth]{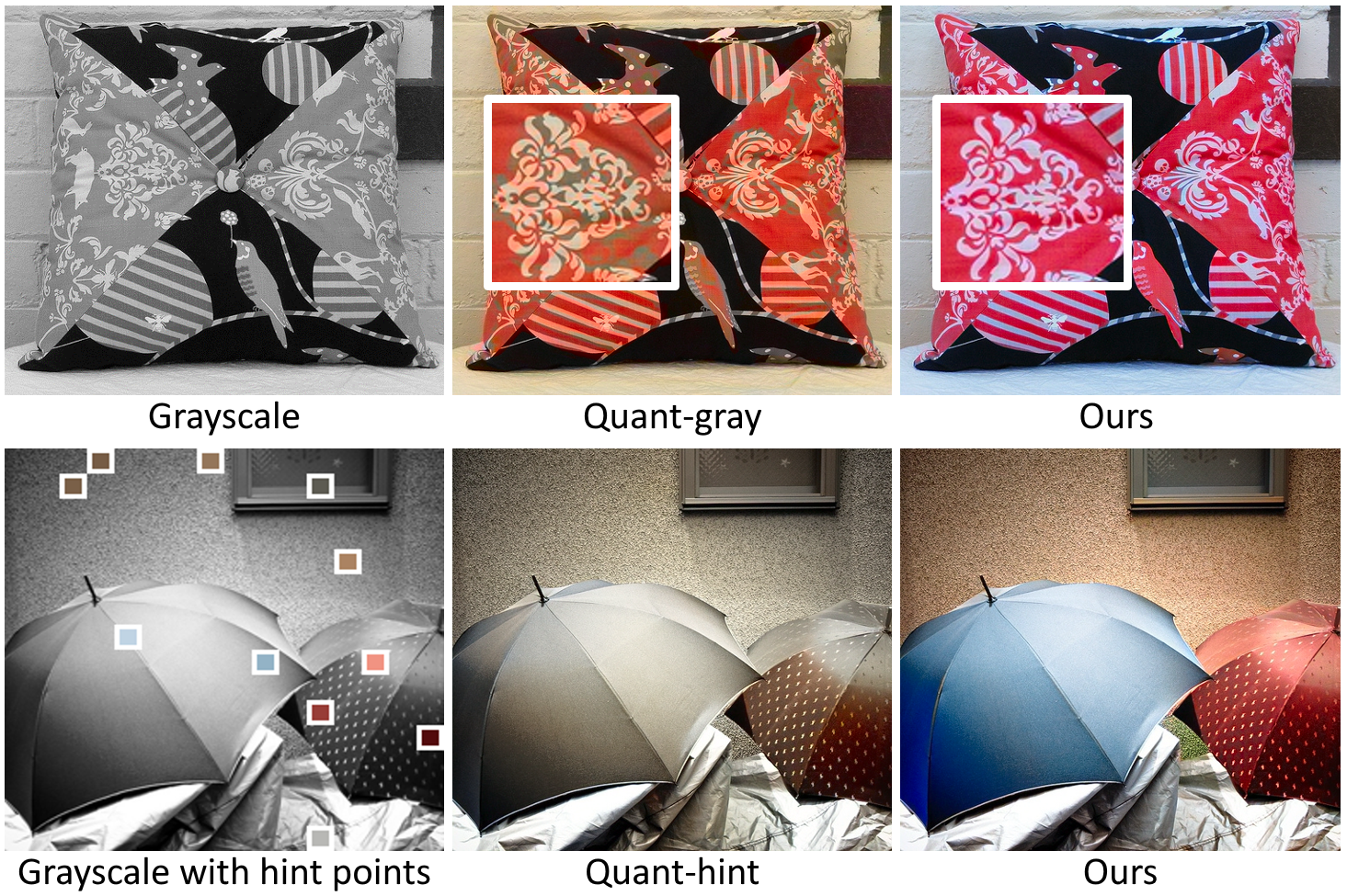}}
	\caption{The effect of Hybrid-Transformer. First row: comparison with Transformer trained using quantized gray features (Quant-gray). Second row: comparison with Transformer trained using quantized hint colors (Quant-hint). \rv{Input images: from ImageNet.} }
	\label{fig:color_comp}
\end{figure}

We show the results in \reftab{ablation_tran} and \reffig{color_comp}. By using continuous gray features, our model outperforms the baseline (Quant-gray) by a large margin in terms of FID and LPIPS and achieves comparable results on Colorfulness. As for the qualitative result ($1^{st} row$ in \reffig{color_comp}), the predicted colors from Quant-gray are not aligned with the input content, leading to a mess of colors. By using continuous hint points, our model outperforms the baseline (Quant-hint) on all three metrics significantly. As shown in the $2^{nd}$ row, the Quant-hint method cannot correctly inherit the color of the hint points, especially when the input hint points are in different colors. The results imply that our Hybrid-Transformer can receive more original and accurate conditional information from the unquantized colors of hint points.

\section{Applications}

With the interactive system and our flexible framework, we show various practical applications in this section, including colorization with hybrid controls, recolorization, and iterative editing. Rather than only showing results on natural images, we also demonstrate more results on legacy old photos. To enable the model to deal with images with diverse contents, we retrain our model on MSCOCO dataset for showing results in this section.

\subsection{Colorization with Hybrid Controls}

\begin{figure*}[t]
	\centerline{\includegraphics[width=1.0\textwidth]{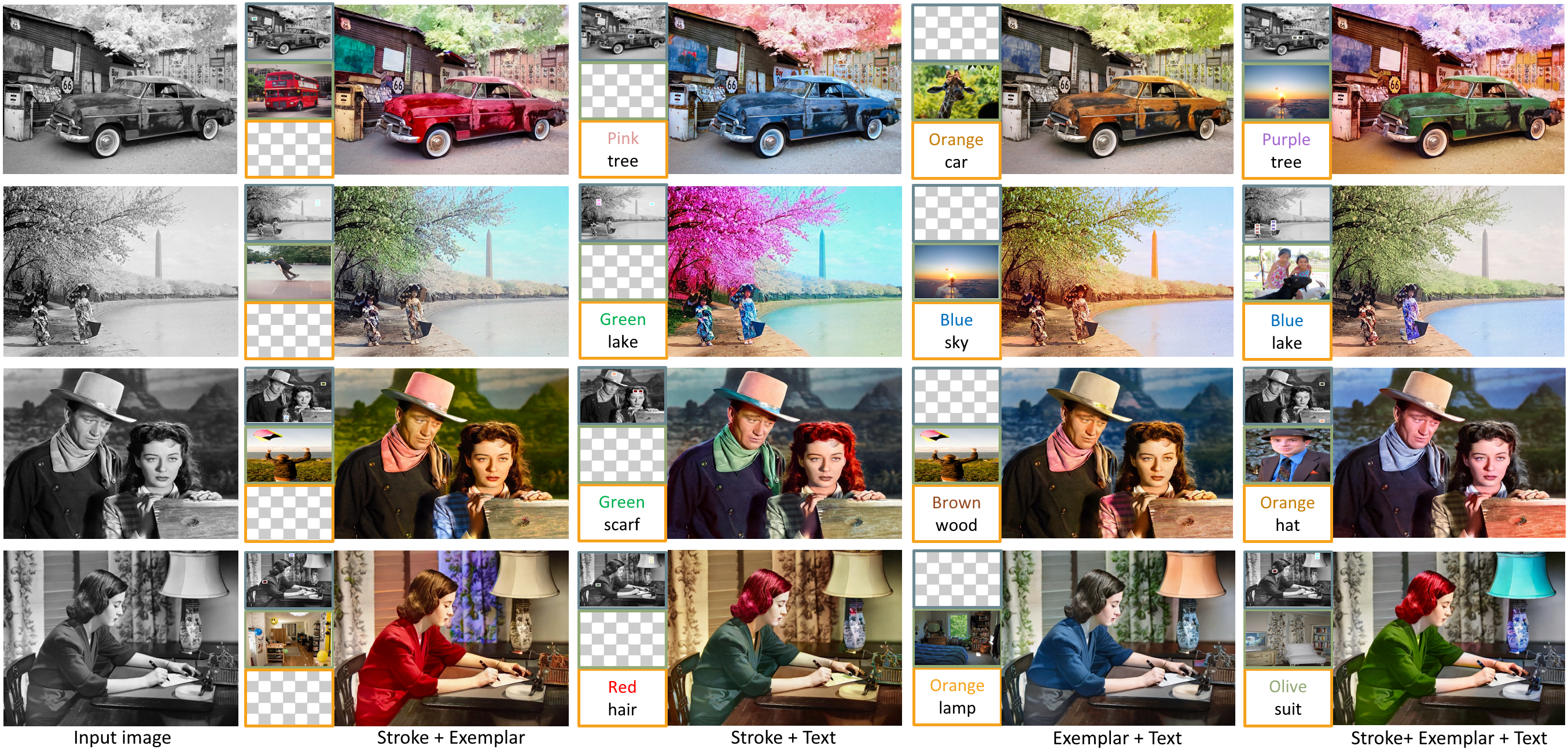}}
	\caption{Colorization on legacy old photos with hybrid controls. In each row, we first show the input grayscale photo, followed by four design cases under different hybrid conditions. We indicate stroke-based conditions with blue boundaries, exemplar-based conditions with green boundaries, and text-based conditions with orange boundaries. \rv{Input images (from top to bottom): 1) Oldtimer automobile Crom; 2) Sumi and Sada Tamura, 1925; 3) Gail Russell \& John Wayne in \textit{Angel and the Badman}, 1947; 4) Portrait of woman writing letter at desk, 1950. All reference images are from MSCOCO.}}
	\label{fig:hybrid_control}
\end{figure*}

Our model allows hybrid-modality for controlling the generated colorization results. We show the results in \reffig{hybrid_control}, and they follow the input hybrid conditions well. For example, in the last column of $4^{th}$ row, we use strokes to control the color of the lamp and hair, reference image to control the color of the curtain, and text for the color of the suit.

\subsection{Recolorization}

\begin{figure*}[t]
	\centerline{\includegraphics[width=1.0\textwidth]{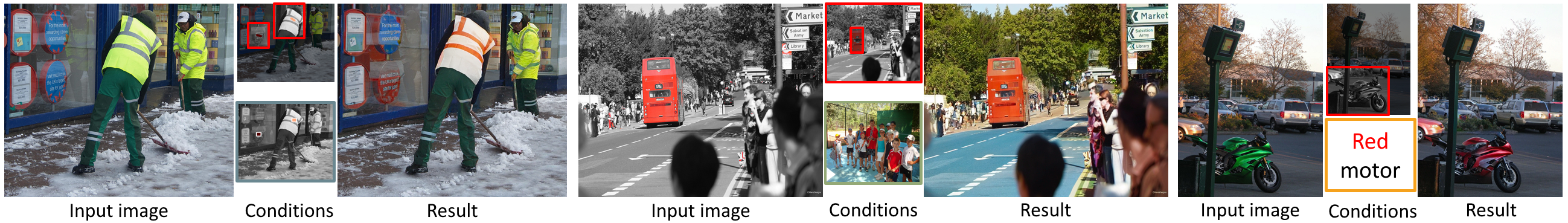}}
	\caption{Recolorization supported by our framework. For each design case, in the middle column, we show the selected region in the upper row, and the input condition in the lower row. \rv{Input images: all images are from MSCOCO, except the first input image is from ImageNet.} }
	\label{fig:recolorize}
\end{figure*}

Except for colorization, our model also allows recolorization of existing color images under different controls. To recolorize the selected region, we simply mask the color tokens within the region and resample them with given conditions. We show the results in \reffig{recolorize}. Our model can adjust the colors of different objects, such as the jacket in $1^{st}$ example (from green to orange) and the motor in the $3^{rd}$ example (from green to red). Besides, instead of only recolorizing on single objects, our model is able to recolor larger scenes, such as the context of the orange bus in the $2^{nd}$ example.

\subsection {Iterative Editing}

\begin{figure*}[t]
	\centerline{\includegraphics[width=1.0\textwidth]{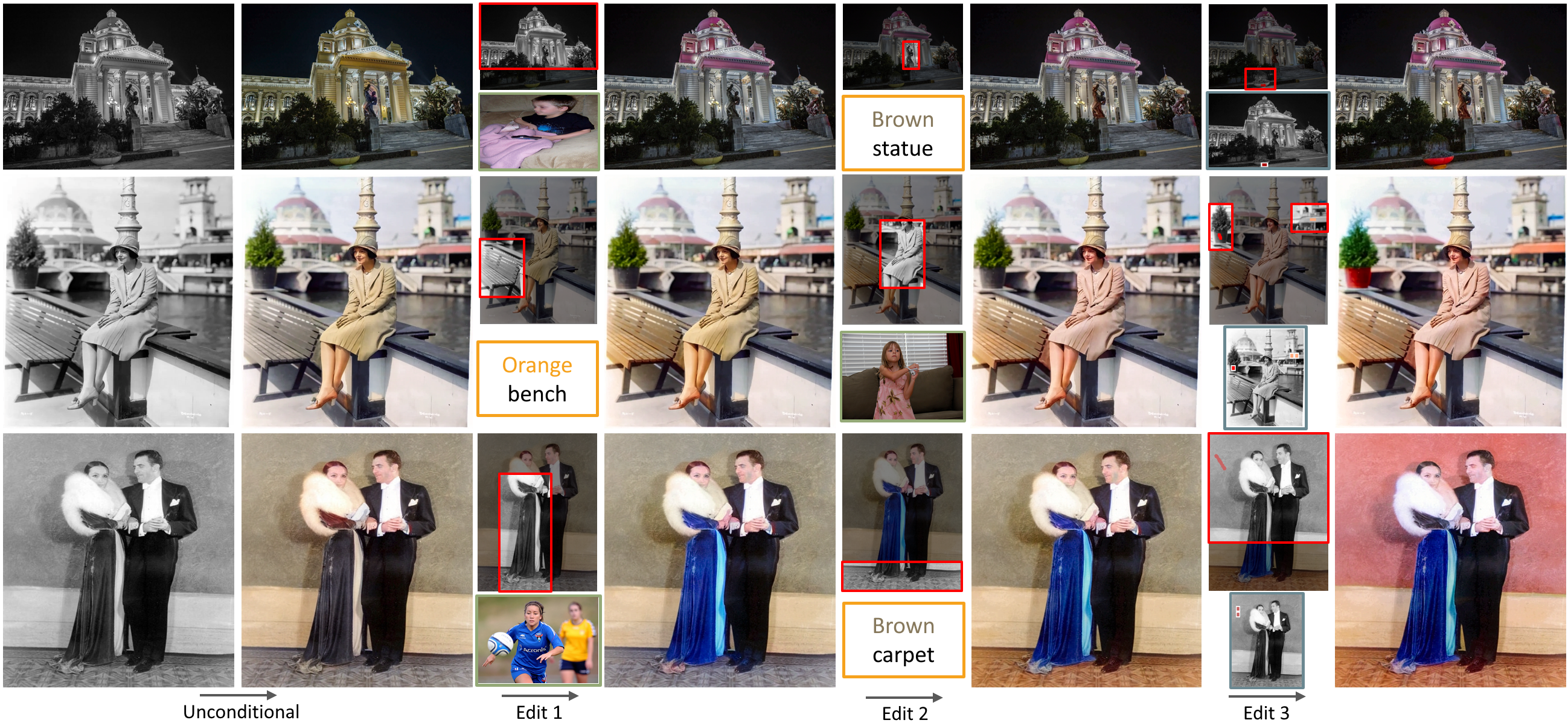}}
	\caption{Iterative editing on legacy old photos. Users can adjust the color based on various conditions iteratively. \rv{Input images (from top to bottom): 1) Night view of the serbian national assembly building; 2) Janet Gaynor, 1920s Coney Island; 3) Dolores del Río next to her husband Cedric Gibbons. Reference images: from MSCOCO.} }
	\label{fig:cont_edit}
\end{figure*}

When colorizing an image, a single pass is often not enough to fully convey the user's intention, and iterative editing becomes an essential way for adjusting and improving the result. Our model is also capable of doing so. We show several results in \reffig{cont_edit}. For tiny objects (\eg~the statue in the $1^{st}$ row and the pot in the $2^{nd}$ row), the users can use strokes or texts to edit the colors. For large objects (\eg~the building in the $1^{st}$ row and the dress in the $3^{rd}$ row), the users can use reference images to control the colors. This also reflects the convenience and flexibility of our system, where users can select different controls to edit various types of objects and images.
\section{Conclusions \& Limitations}

To conclude, we propose the first unified framework UniColor, which supports diverse colorization in different modalities, including both unconditional and conditional ones (\eg~stroke, exemplar, and text). Different from existing works, which only support a specific type of user control and cannot generalize to other ones, our framework unifies all three types of controls into the form of hint points, which can be naturally extracted from stroke and exemplar conditions. To extract hint points from the text input, we propose a novel CLIP-based method to locate the objects described in the text and add corresponding colors to form the hint points. With the hint points, we propose a network, which consists of a Chroma-VQGAN and Hybrid-Transformer, for diverse colorization with high quality and colorfulness. Based on our unified framework, we design an interactive system to support all four types of image colorization (\ie~unconditional, stroke-based, exemplar-based, and text-based). The system also enables the user to perform hybrid controls, re-colorizing, and iterative editing. 

\begin{figure}[t]
	\centerline{\includegraphics[width=1.0\linewidth]{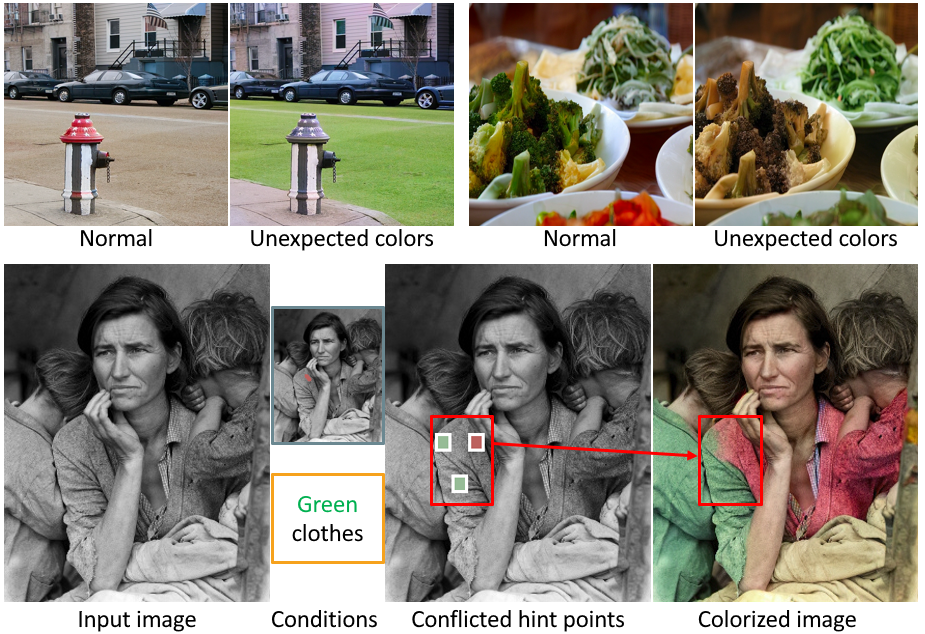}}
	\caption{\rv{Failure cases. $1^{st}$ row: unexpected colors, input images: from MSCOCO. $2^{nd}$ row: conflicts in hybrid controls, input image: Migrant Mother, 1936.} }
	\label{fig:failure}
\end{figure}

Despite the superior performance of UniColor, we still encounter some limitations to be explored in the future. First, diversity is generally an advantage in image colorization, but, on the other hand, the stochastic sampling may seldom lead to unexpected colors, such as the green road and brown broccoli in \reffig{failure} $1^{st}$ row. This can be alleviated by limiting the range of stochastic sampling but sacrificing diversity to some degree. Or we may consider filtering the colorization results based on some semantic metric. For example, we may train a network to evaluate the quality of the produced colors. Moreover, control conflicts might occur when mixing conditions from different modalities. As shown in \reffig{failure} $2^{nd}$ row, the green hint points from the text condition, and the red stroke lay on the same clothes and thus generated a mixture of green and red. This problem can be avoided by more careful user interactions or designing an algorithm to auto-detect the conflicts based on image segmentation. For example, if the algorithm detects that the hint points from different controls lay on the same segment, it will automatically ignore some hint points according to the default priority or notify the user to make further decisions. It would be worthwhile to further improve user experiences in colorization tasks with our Unicolor framework.

\section{Acknowledgments}

We thank the anonymous reviewers for helping us to improve this
paper.
We also thank the artists and photographers for approving us to use their photos.
This work was supported by the Hong Kong Research Grants Council (RGC) GRF Scheme under Grant CityU 11216122.

\bibliographystyle{ACM-Reference-Format}
\bibliography{reference}

\end{document}